\documentclass[twoside,11pt]{article}

\usepackage{jmlr2e}
\usepackage{amsmath}
\usepackage[format=hang]{subcaption}

\newcommand{\bbeta}{\boldsymbol{\beta}} 
\newcommand{\bepsilon}{\boldsymbol{\epsilon}} 
\newcommand{\btheta}{\boldsymbol{\theta}} 

\newcommand{\bxi}{\boldsymbol{\xi}}
\newcommand{\bsigma}{\boldsymbol{\sigma}}
\newcommand{\bnu}{\boldsymbol{\nu}}

\newcommand{\bSigma}{\boldsymbol{\Sigma}}

\newcommand{\sB}{\mathcal{B}}

\newcommand{\sN}{\mathcal{N}}
\newcommand{\sP}{\mathcal{P}}

\newcommand{\sY}{\mathcal{Y}}

\newcommand{\bB}{\mathbf{B}}
\newcommand{\bC}{\mathbf{C}}
\newcommand{\bD}{\mathbf{D}}

\newcommand{\bI}{\mathbf{I}}

\newcommand{\bW}{\mathbf{W}}

\newcommand{\bY}{\mathbf{Y}}

\newcommand{\bd}{\mathbf{d}}

\newcommand{\bx}{\mathbf{x}}
\newcommand{\by}{\mathbf{y}}
\newcommand{\bz}{\mathbf{z}}

\newcommand{\bone}{\mathbf{1}}
\newcommand{\bzero}{\mathbf{0}}

\DeclareMathOperator{\E}{\mathbb{E}}

\DeclareMathOperator*{\Cov}{\mathrm{Cov}}

\DeclareMathOperator{\Diag}{Diag}
\DeclareMathOperator{\R}{\mathbb{R}}

\newcommand{\bp}{\boldsymbol{p}}
\newcommand{\balpha}{\boldsymbol\alpha}

\usepackage{tikz}
\usetikzlibrary{bayesnet,matrix}

\pgfdeclarelayer{edgelayer}
\pgfdeclarelayer{nodelayer}
\pgfsetlayers{edgelayer,nodelayer,main}

\definecolor{hexcolor0xbfbfbf}{rgb}{0.749,0.749,0.749}

\tikzset{>=latex}
\tikzstyle{none}   = [inner sep=0pt]

\tikzstyle{line}  = [ - ]
\tikzstyle{arrow}  = [ ->, shorten <=1pt, shorten >=1pt ]
\tikzstyle{ardash} = [ dotted, ->, shorten <=1pt, shorten >=1pt ]

\tikzstyle{empty}=[circle,opacity=0.0,text opacity=1.0,inner sep=0pt,minimum
width=0pt,minimum height=0pt]
\tikzstyle{box}=[rectangle,fill=White,draw=Black]
\tikzstyle{filled}=[circle,fill=hexcolor0xbfbfbf,draw=Black]
\tikzstyle{hollow}=[circle,fill=White,draw=Black]
\tikzstyle{param}=[rectangle,fill=Black,draw=Black,inner sep=0pt,minimum width=4pt,minimum height=4pt]

\ShortHeadings{A Contemporary Overview of Probabilistic Latent Variable Models}{Farouni}
\firstpageno{1}

\begin{document}

\title{A Contemporary Overview of Probabilistic Latent Variable Models}

\author{\name Rick Farouni \email tfarouni@mgh.harvard.edu \\
       \addr Department of Molecular Pathology,\\
        Massachusetts General Hospital,\\
         Harvard Medical School}

\editor{}

\maketitle

\begin{abstract}
In this paper we provide a conceptual overview of latent variable models within a probabilistic modeling framework, an overview that emphasizes the compositional nature and the interconnectedness of the seemingly disparate models commonly encountered in statistical practice. 
\end{abstract}


\section{Introduction}

In this paper, we will be concerned with building probabilistic latent variable models for data that populate an $N \times P$ matrix $\mathbf{Y}$ whose every element $y_{np}\in \R$ is real-valued measurement.  
\begin{align}\label{matrix}
\mathbf{Y}=\begin{bmatrix}
\by_1^{\rm T} \\
\vdots \\
\by_n^{\rm T}\\
\vdots \\
\by_N^{\rm T}
\end{bmatrix} 
=\begin{bmatrix}
    y_{1,1}  & \dots  & y_{1,P} \\
    \vdots  & \ddots & \vdots \\
    y_{n,1}  & \dots  & y_{n,P} \\
    \vdots  & \ddots & \vdots  \\
    y_{N,1}  & \dots  & y_{N,P}
\end{bmatrix} 
\end{align}

In general, we will be limiting our focus mainly to the the multivariate statistics setting in which we treat an observation as a multivariate random vector $\by_n = \left(y_{n,1}, \hdots,y_{n,P}\right)$ consisting of $P$ features and the data as a set of $N$ observations $\by=\{\by_1, \cdots, \by_n, \cdots, \by_N \}$. Accordingly, we can think of  the data matrix  $\mathbf{Y}$ as consisting of $N$ observations, each a $P$-dimensional vector, that populate the rows of the matrix.

\section{Basic Definitions}\label{sect:statmodel} 

In order to give a rigorous constructive account of probabilistic latent variable models, we begin with defining several foundational concepts that will later serve as our building blocks.  In what follows, we attempt to convey the intuition at the expense of mathematical rigor. More rigorous  definitions can be found in \citet{durrett2010} and \citet{athreya2006}.
  
\subsection{What is a Probability Model?} 

Our first building block is a probability model. A \emph{probability model} is defined as the triplet consisting of a set, a set of sets, and a function. More precisely given by
\[ 
\left(\mathcal Y , \ \mathcal B(\mathcal Y),\ p\right)
\] 
where  
\begin{enumerate}
\item  $\mathcal Y$ is the \emph{sample space} 
\item  $\mathcal B\left(\mathcal Y \right)$ is the \emph{Borel $\sigma$-algebra} 
\item $p$ is a \emph{probability measure} 
\end{enumerate}

Whereas the sample space is a relatively easy concept to understand - it is a set of all possible \emph{outcomes} - the notion of a Borel $\sigma$-algebra tends to be more involved. So what exactly is a Borel $\sigma$-algebra? The short answer is that it is the set of all \emph{events} of interest; technically, the set of subsets of $\mathcal Y$. For the long answer, we first need to define an \emph{algebra}. An algebra in this context is a set along with a collection of operations. Here is an example. First take our sample space, the set $\mathcal Y$, and equip it with two binary operations $\left(\cup,\cap\right)$, the union and intersection. Now let us consider two events only - the empty set $\emptyset$  denoting the event that nothing happens and the sample space $\mathcal Y$ denoting the event that all outcomes enumerated in the sample space occur. Starting with these two events only, we can, using the two operations, construct the set of all possible subsets of $\mathcal Y$, which is called \textit{the powerset}
\[ 
\big(\mathcal P\left(\mathcal Y\right),\cap,\cup,\neg,\emptyset,\mathcal Y\big)
\] 
Since every possible subset of $\mathcal Y$ is an event, accordingly the powerset consists of all possible events. With a finite number of outcomes, the number of all possible events is finite as well. Therefore, the powerset is an algebra that is closed under the union and intersection of finitely many subsets. However, when our sample space is not finite, but rather countably infinite, we would instead need a specialized set of subsets to enumerate all possible events. That set is called a $\sigma$-algebra. It is simply a sub-algebra of the powerset, completed to include countably infinite operations. Lastly, when the events are subsets of the real line $\mathbb{R}$, we would need to resort to a more specialized algebra, called the Borel $\sigma$-algebra. It is defined as the $\sigma$-algebra generated by the open sets and which can be constructed from open or closed sets by repeatedly taking countable unions and intersections. In notation, we define it as such. 
\[ 
 \mathcal B(\mathbb{R}):=\sigma(\mathcal C); \quad \mathcal C= \lbrace(a,b],-\infty\leq a\leq b\leq\infty\rbrace
\] 
Whereas the powerset is the largest possible $\sigma$-algebra, the collection of subsets $\lbrace  \emptyset,\mathcal Y \rbrace$ is the smallest. The Borel $\sigma$-algebra lies between the two extremes
\[ 
\lbrace  \emptyset,\mathcal Y \rbrace \subset  \mathcal B(\mathcal Y) \subset \mathcal P(\mathcal Y)
\]
and it is the smallest $\sigma$-algebra containing all open sets in $\R$ (hyper-rectangles in $\R^{P}$). 

The third component of a probability model involves the notion of a measure. What is a probability measure? It is simply a function $\mathit p:\mathcal B\left(\mathcal Y\right) \to [0 ,1]$ that maps events to a real valued number between zero and one, thus assigning each event a finite probability of occurring. The notion of a probability measure is tightly intertwined with the concept of a random variable. To illustrate, our multivariate observation $\by= \left(y_1, \hdots,y_P\right)$ is a real-valued continuous random vector with distribution $\mu$ that takes values in the \emph{measurable space} $\big(\mathcal Y,\ \mathcal B(\mathcal Y)\big)$. Given the probability triplet, we can define a random vector (with a slight abuse of notation) as a measurable function from the sample space to a $P$-dimensional real space  $\by: \mathcal Y \mapsto \R^P$ such that
\[ 
\lbrace  \by \leq \textrm{y}  \rbrace \in B; \ where \quad B \in \mathcal B \left(\R^P \right)
\] 
The important point is that the distribution of $\by$ induces a probability measure on the measurable space $\big(\mathcal Y,\ \mathcal B(\mathcal Y)\big)$ and is given by
\[ 
\mu(B) = p\left( \by \in B\right); \ where \quad B \in \mathcal B \left(\R^P \right)
\] 
where $p\left( \by \in B\right)$ denotes the probability that the event $\lbrace  \by \leq \textrm{y}  \rbrace$ lies in a Borel set $B$. Note that a \emph{Borel set} $B$ is an element of a Borel $\sigma$-algebra.

\subsection{What is a Statistical Model?} 

We build a probabilistic model starting with the joint distribution of all observable and unobservable variables. Since we have not introduced any unobservable quantities yet, our entry point is the following joint probability distribution for all $N$ observations,
\begin{align}\label{eq:probmodel} 
p\left(\by\right)= p\left(\by_1, \by_2, \cdots, \by_N \right)
\end{align} 
where $p$ is a joint probability density function corresponding to a specific, known probability measure, or equivalently, to a known random variable.  

Technically, the joint density given by Equation \ref{eq:probmodel} denotes a probability model, not a statistical model. To illustrate, let $\mathcal{M}(\mathcal Y)$ denote the \textit{space of all probability measures} on the measurable space $\big(\mathcal Y,\ \mathcal B(\mathcal Y)\big)$. As can be seen, there are infinitely many possible probability measures we can choose from. That is, $\mathcal{M}(\mathcal Y)$ can be too large a space for many inference problems with limited data. To make inference a less daunting task we can restrict ourselves to a indexed subset of the space by introducing an unobservable variable, a \textit{parameter} $\theta \in \Theta$ \ that represents the pattern that explains the data, where the \textit{parameter space} $\Theta$ is the set of all possible values of $\theta$.  As a result, the probability measures $P_\theta$ are now elements of $\mathcal{PM}(\mathcal Y)$, the space of all probability measures on $\Theta$ with elements $P_\theta \in\mathcal{PM}(\mathcal Y)$ indexed by a parameter $\theta \in \Theta$.  

Accordingly, a \textit{statistical model} $ \mathcal P$ can be defined as a subset of the space of all probability measures $\mathcal P \subset \mathcal{M}(\mathcal Y)$ such that $ \mathcal P=\lbrace P_\theta \mid \theta \in \Theta\rbrace$ where $\theta \to P_\theta$ is a bijective and measurable assignment. We say that the model $ \mathcal P$ is \textit{parametric statistical model} if $\Theta \subset \mathbb{R}^d$ for some $d \in \mathbb{N}$. The subset $ \mathcal P \subset \mathcal{PM}(\mathcal Y)$ can be even restricted further. We can specify a \textit{family of parametric models} $ \mathcal G=\lbrace G_\theta \mid \theta \in \Theta\rbrace$ where $\theta \to G_\theta$ is smooth. For example, $\mathcal G=\lbrace N(\theta,1): \theta \in \Theta\rbrace$ specifies the one-dimensional normal location family of models. If on the other hand we choose the parameter space $\Theta$  to be infinite dimensional, then we say that $ \mathcal P$ is \textit{a nonparametric statistical model} such that $\Theta$ is equivalent to $\mathcal{PM}(\mathcal Y)$, the space of all probability measures on $\Big(\mathcal Y, \mathcal B(\mathcal Y)\Big)$. As a result, we obtain the following hierarchy of probability model spaces.
\[ 
\mathcal G \subset  \mathcal P \subseteq \mathcal{PM}\subset \mathcal{M}
\]
Now, by introducing the  parameter $\btheta$,  we can now specify the joint density of a parametric statistical model for our multivariate observations as such
\begin{align}\label{eq:statmodel} 
p\left(\by ; \btheta \right)= p \left(\by_1, \by_2, \cdots, \by_N; \btheta \right)
\end{align} 
where $\btheta \in \Theta \subset \mathbb{R}^d$. In the case of a parametric statistical model, the superscript $d$ is a finite integer, whereas  in the case of a nonparametric statistical model, it is an infinite integer. Note that we use the semicolon to denote that the parameter $\btheta$ is not a random variable.

\subsection{What is a Bayesian Model?} 

In frequentist statistics, we treat a parameter $\btheta$ as an unobservable unknown quantity and estimation involves finding $\btheta^\ast \in \Theta$ that governs the distribution of the data we observe if the true generating model indeed lies inside the family of probability models we specify. In the probabilistic perspective, we treat a parameter as an unobservable random quantity whose prior distributions quantifies our initial uncertainty. Learning involves computing the posterior distribution of $\btheta$ that quantifies the residual uncertainty that results after we observe the data. 

Accordingly, the joint probability density of all observable and unobservable random variables can now be given by
\begin{align}\label{eq:probalistic_model} 
p\left(\by , \btheta \right)= p\left(\by_1, \by_2, \cdots, \by_N, \btheta \right)
\end{align} 

Note that Equation \ref{eq:probalistic_model} defines a general probabilistic model, not a Bayesian model. To reduce it to a Bayesian model, we need to factorize the joint into the product of a conditional data distribution and a marginal distribution prior distribution as such.
\begin{align}\label{eq:bayesianmodel2} 
p\left(\by , \theta \right)=  \prod_{n=1}^N p\left( \by_n \mid \btheta\right) p\left( \btheta \right)
\end{align}

This factorization entails a particular set of assumptions. To illustrate them, we need to define a Bayesian model first.

A \textit{parametric Bayesian statistical model} ($\sP, \Pi$) consists of a model $\sP$, the \textit{data distribution}, and a \textit{prior distribution} $\Pi$ on $\Theta$ such that $\btheta$ is a random variable taking values in $\Theta$ given that $\Pi\Big( \{ P_{\btheta}: \btheta \in \Theta \}\Big)=1$. A Bayesian model is a generative model in which the data is generated hierarchically in two stages:
\begin{align}\label{eq:parm_bayesian_model}
\begin{split}
\btheta &\sim \Pi  \\
\by_n \mid  \btheta  & \overset{i.i.d.}{\sim} P_{\btheta}  \qquad \btheta \in \Theta \subset \mathbb{R}^d 
\end{split}
\end{align}

A Bayesian model can be thought of as a \textit{random mixture model} where we first sample from a mixing measure $\btheta \sim \Pi$, then sample from a component $\by_n \mid  \btheta  \sim P_{\btheta}$. After we observe the data, we then update the prior to the posterior $\Pi(.\mid \by_1,\by_2,\cdots)$. 

In contrast, a \textit{nonparametric Bayesian model} is a Bayesian model whose prior $\Pi$ is defined on an infinite dimensional parameter space $\Theta$. The corresponding two-stage hierarchical model can be given as such.
\begin{align}\label{eq:nonparm_bayesian_model}
	\begin{split}
P &\sim \Pi  \\
\by_n \mid P  &\sim_{iid} P \qquad P \in \sP
    \end{split}
\end{align}
A prior distribution on an infinite dimensional space is a \textit{stochastic process}. Defining an infinite dimensional prior distributions is not straightforward, but one way to construct a prior distribution $\Pi$ on $\Theta$ is through De Finetti's Theorem.

\begin{theorem}[De Finetti's Theorem]
A sequence of random variables $\{\by_n\}_{n=1}^\infty$ with values on $\sY$ is \textit{exchangeable} if and only if there is a unique measure $\Pi$ on $\Theta$ such that for all $N$
\begin{align}
p(\by_1,\by_2,...,\by_N) = \int_{\Theta}\left( \prod_{n=1}^N p(\by_n\mid \theta)\right)p(\theta) d\theta
\end{align}
\end{theorem}
De Finetti's gives us a infinite mixture representation of the joint probability of the observations \citep{Ghosh2006}. More importantly, the theorem implies that there exists a random probability measure $P$ such that the sequence of exchangeable observations are $i.i.d$ conditional on $P$. Implicit in the resulting mixture representation is a data-generating process that gives rise to the observations.

What is exchangeability? We say a sequence of random variables is exchangeable when their distribution is invariant under permutation of the indices.  When building statistical models, we can use exchangeability and De Finetti's Theorem to treat an infinite exchangeable sequence as a conditionally $i.i.d$ sequence of random variables. In data applications, exchangeability implies that the future is like the past. 

\paragraph{Distributional Symmetries}  It is important to highlight here that exchangeability is one of the four basic probabilistic symmetries defined by \citet{kallenberg2006}; namely, stationarity, contractability, exchangeability, and rotatability. These four distributional symmetries correspond to invariance properties under the four respective transformations of shifts, sub-sequences, permutations, and isometries. The four probabilistic symmetries form a hierarchy with stationarity at the bottom and rotatability, the strongest invariance property, at the top. Exchangeability in particular allows us, through conditioning, to transform the joint density into a mixture of densities.

\paragraph*{Example of a Bayesian model} When $\mathcal Y$ is finite, there is usually a natural unique measure $\Pi$ we can obtain. More specifically, if $\mathcal Y=\{1,2,\ldots,K\}$, then $\mathcal{PM}(\mathcal Y)=\{(p_1,\ldots,p_{K}): 0\leq p_k\leq1 ,\sum p_k=1 \}$. That is, the space of probability measures corresponds to a simplex parametrized by a $K-1$ dimensional vector $\bp=(p_1,\ldots,p_{K-1})$. A natural prior $\Pi$ to specify on $\bp$ is the Dirichlet distribution. For example, consider the Bayesian model($\mathcal P, \Pi$) where the observation model $\mathcal P$ is the \emph{Categorical distribution} defined on the sample space $\mathcal Y=\{1,2,\ldots,K\}$, and the prior $\Pi$ is the \emph{Dirichlet distribution} defined on the simplex $\Theta=\{(p_1,\ldots,p_{K}): 0\leq p_k\leq1 ,\sum p_k=1 \}$
\begin{align*}
\bp  &\sim \operatorname{Dirichlet}( \boldsymbol\alpha) \\
y_i\mid \bp &\sim \operatorname{Categorical}(\bp)\\
\textit{where} \quad\boldsymbol\alpha &= \left(\frac{\alpha}{K}, \cdots, \frac{\alpha}{K} \right) \\
\bp &= \left(p_1, \cdots, p_K \right)  
\end{align*}

\begin{figure}[htb]
    \includegraphics[width=.8\textwidth]{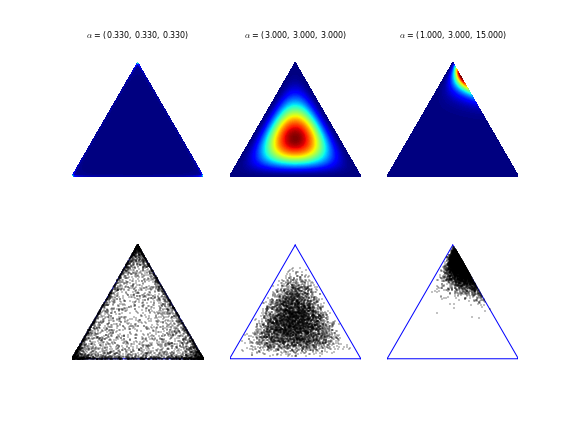}
    \captionsetup{skip=2pt}
  \caption{Dirichlet Distributions}
  \label{Dirichlet}
\end{figure}

Here, \textit{the Dirichlet distribution} is the conjugate prior of the Categorical distribution and the concentration hyperparameter vector $\balpha$ represents the number of pseudo-observations, the \textit{a priori} weights, for each of the $K$ clusters. Figure \ref{Dirichlet} shows density and sampling plots corresponding to Dirichlet distributions for $K=3$ and several choices of $\balpha$.

The Dirichlet distribution is a conjugate prior that allows us to sample $iid$ from  $\mathcal Y=\{1,2,\ldots,K\}$. To generalize, we can make $\mathcal Y=\R$ thus obtaining the space of all measures $\mathcal{M}(\R)$ defined on the measurable space $(\R,\sB)$, which is the real line equipped with the \textit{Borel} $\sigma$\textit{-algebra}. It can be shown \citep{Ghosh2006} that for every partition $B_1,B_2,\ldots,B_k$ of the real line $\R$ by Borel sets, there exists a unique measure $DP_\alpha$ on $\mathcal{M}(\R)$ called the Dirichlet Process with parameter $\alpha$ satisfying 
\[
(P(B_1),P(B_2),\ldots,P(B_k)) = DP(\alpha B_1,\alpha B_2,\ldots,\alpha B_k))
\]
That is, when $\mathcal Y=\R$, then $\mathcal{M}(\R)$, is the space of all probability measures on $\R$. If the sample space $\mathcal Y=\R$ is partitioned into measurable subsets, then for every partition $ (B_1, B_2,\ldots, B_k)$, the prior probability measure $\Pi$ on  $( p(B_1), p(B_2),\ldots, p(B_k))$ is a Dirichlet process prior. 

The Dirichlet process $DP(G_0,\alpha)$ is the simplest distribution - in terms of the extent of independence it assumes - that can have this form. That is probably why it has been referred to as \emph{the normal distribution of Bayesian nonparametrics}. More specifically, the location of the atoms, the $\theta's$, are sampled $iid$ but the weights are sampled as independent proportions. Now, the weights cannot be sampled $iid$ because then they will not sum up to one. However, the next best thing to $iid$ sampling is to obtain independent proportions, as is done using the stick-breaking representation.

\section{The Probabilistic Approach to Latent Variable Models} 

We begin by describing the probabilistic modeling approach. Next we explain what a latent variable is and provide a general formulation of the latent variable models. We end the section by giving a brief history of latent variable modeling.

\subsection{The Probabilistic Modeling Approach} 

A quick insight into the essential aspect of the probabilistic modeling perspective can be gained by contrasting it with what \emph{it is not}. The probabilistic approach does not construct statistical models using algebraic relations. In particular, reasoning about how observed and unobserved quantities relate to each other and how their dependencies are induced does not involve specifying structural relationships using algebraic formulas. Instead, all observed and unobserved events that are thought to represent the phenomenon of interest are denoted by random variables and each event is allocated a finite probability of occurring. That is to say, the generative probabilistic modeling approach is Bayesian in spirit \citep{pearl2014, lauritzen1996, jebara2012}. In the probabilistic approach, we start with the joint distribution of all observed and unobserved quantities that govern the distribution of the data.
\[
p \left(\by, \btheta \right)
\]
We construct a probabilistic model in two steps. In the first step, we factorize the joint density into a product product of conditional and marginals distributions. Through conditioning, we introduce dependencies that transform the joint density into a mixture of densities. Implicit in this mixture is a data-generating process that gives rise to the observations. As Joe Blitzstein has often said, conditioning is the soul of statistics. Indeed, it is the most important tool in a statistician's toolkit for it can make inference more tractable by allowing us to factorize any joint distribution into the product of several conditional probability distributions. In the second step, we specify particular distributions to the factorized densities. 

The generative approach requires much more thought to apply in practice since there are a large number of ways we can decompose the joint distribution into a product of factors. Since each possible decomposition implicitly encodes a particular number of modeling assumptions, the researcher is forced to examine the validity of and the motivation for all modeling assumptions. Furthermore, each component of the factorization needs to be assigned a probability distribution that respects the plausibility of what is known about the putative mechanism that gives rise to the data. For instance, in order to treat the measurements as random in the first place, we would need to first quantify the sources of uncertainty in inherent in the randomness using a statistical model. Randomness can arise from either \emph{experimental error}, \emph{model inadequacy}, \emph{parametric variability} induced by random sampling from a population, or even \emph{parameter uncertainty} arising from our inability to exactly know the true values of the parameters \citep{Kennedy2001}. 

However, before we start making assumptions about the structure of the model or the distributional form of its component factors, we need to examine two important questions first; namely, how the data was collected, and what should its grouping structure be.  We start with the first issue.  

\subsubsection{Recording Measurements in Space and Time} 

In the probabilistic modeling framework, we begin the task of building a statistical model, not with a particular model in mind, but rather with the data. Consider once again the data matrix
 
\begin{align}\label{matrix}
\mathbf{Y}
=\begin{bmatrix}
    y_{1,1}  & \dots  & y_{1,P} \\
    \vdots  & \ddots & \vdots \\
    y_{n,1}  & \dots  & y_{n,P} \\
    \vdots  & \ddots & \vdots  \\
    y_{N,1}  & \dots  & y_{N,P}
\end{bmatrix} 
\end{align}

We approach the task of building a statistical model for the data by first examining the individual measurement events that produce the recorded observations. Since the event of taking a measurement occurs within a four-dimensional spatiotemporal frame of reference, we denote a single measurement not as $y_{np}$, but rather as $y(d_1,d_2,d_3,t)$, a measurement recorded at spatial location $\bd=(d_1,d_2,d_3)$ and time $t$, where $\bd \in D \subset \R^3$ and  $t \in T \subset \R$. Note that we do not consider here point process models in which $D$ is random such that $\bd$ denotes the location of random events. 

In many applications, spatiotemporal information is assumed to have a miniminal effect on the observed data and is therefore omitted. For example, if we wish to measure the heights of a number of individuals at a certain point in time, it could be safe to assume that the spatial arrangement of the subjects and the brief time interval that passes between the measurements are inconsequential. If on the other hand, the height measurements for each subject are taken multiple times over a period of months, then the time index would need to be retained. To illustrate, if the height of individual $i$ is recorded repeatedly over three occasions, we would then have the measurements $[y(t^*_1), y(t^*_2), y(t^*_3)]$. For uniform time intervals, the notation can be simplified and the vector of repeated measures can be denoted as $[y_1, y_2, y_3]$. In other applications, the location of measurements is an essential aspect of understanding the phenomenon and cannot be ignored.

At any rate, the form in which the data is presented implicitly imposes certain assumptions on us. For example, if the data  come in the form of a data matrix without an accompanying spatiotemporal index for each measured response  $y_{np}$, then we might incorrectly assume that the measurements were uniformly sampled both in time or space. When the spatiotemporal indices are absent such is the case for the data matrix  $\mathbf{Y}$ given in Equation \ref{matrix}, it would be good practice to consider the following few questions before proceeding to model the data.
\begin{itemize}  
 \item  Are the $P$ measurements that make up a given observation $\by_n$ collected at the same time? If not, are the time lags between them inconsequential?
 \item  Do the spatial indices $\bd$ vary continuously over the three dimensional frame of reference $D$ or is the space $D$ discretized?
 \item Does the time index $t$ vary continuously over its domain $T$ or is the time domain $T$ discretized? 
\item In case of discrete time, are we dealing with multivariate time series (i.e. the spatial coordinates are the same for each time period) or cross-sectional data (i.e. the spatial coordinates vary over the time periods)?     
 \end{itemize} 
 
In what follows, we will be assuming, when applicable, that both the temporal domain $T$ and spatial domain $D$ are discretized uniformly; more specifically, that $D$ is divided into a regular $P$ dimensional lattice where $\bd_p$ represents the spatial coordinates of the $p$th block and that $T$ is divided into equal time intervals, where each measurement represent an average over an time interval and a spatial block. We will also be assuming that the spatial coordinates are fixed for all the time periods and that there is no lag between the $P$ measurements. Consequently, the time and space indices would no longer be needed since the row and column indices of the data matrix are sufficient to preserve the spatiotemporal structure of the measurements, in the sense that the complex 3D pattern of spatial dependencies can still be inferred from the one dimensional response vector.

\subsubsection{The Grouping Structure of the Data} 

In statistical studies, the observed data is usually characterized by a grouping structure that determines what the observational units are. An observation could be either a univariate random variable $y_{n} \in \R$, a multivariate random vector $\by_n = (y_{n1}, \hdots,y_{nP}) \in \R^P$, or even a function  $f_n \in L^2 \subset\R^\infty$, where $f_n$ in this case is a point in the Lebesgue space $L^2$, an infinite dimensional function space. Although the grouping structure is usually determined by the experimental design or by how the data was collected, the statistical model we specify for the data can either ignore, relax, or even impose a particular grouping structure on the data. Our choice of what constitutes an observational unit determines whether we approach the modeling endeavor from a univariate, a multivariate, or a functional data analysis perspective.

To illustrate, let us begin with the joint probability distribution of all  $K \left(=N \times P \right) $ variables that populate matrix $\mathbf{Y}$
\[ 
p\left(y_{11}, \hdots, y_{np}, \hdots, y_{NP} , \btheta\right) 
\] 

If we map the two-subscript index set to a single -subscript index  set, we can express the joint as such
\[ 
p\left(y_1, \hdots, y_k, \hdots, y_K , \btheta\right) 
\]

Now if we refrain from making any distributional symmetry assumptions regarding the $K$ random variables, we then have to treat all $K \ $ measurements as a \emph{single} high-dimensional sample from a joint distribution $p_{\btheta}$ governed by parameter $\btheta$. That is, since we do not assume that the joint distribution of the variables is invariant to permutation or any other transformation of the indices, our data would  effectively consist of one single $K$ dimensional observation sampled from a distribution $p_{\btheta}$
\[ 
\left(y_1, \hdots, y_k, \hdots, y_K \mid \btheta \right) \sim \quad p_{\btheta}  
\]

Such a model would be plausible if our single observation is a matrix-valued random variable  $\bY \in \R^{N\times P}$ such that of an image or a two-dimensional spatial data observation. In principle, the joint model can account for any complex pattern of dependencies among the $K$  measurements irrespective of their row or column index. Nonetheless, the joint distribution of a large number of random variables would require a large number of parameters to enable it to capture the dependencies among all possible combination of the variables. Furthermore, given the complex form of dependencies that could be present in the data, statistical inference would be a hopeless task with only one single sample. Note that in general the parameter $\btheta$ can be a scalar, a multidimensional vector, or even  an infinite dimensional vector function. However, since for any scalar observation, no more than one parameter can be learned, we are effectively limited by the number of parameters we can learn given the $K$ measurements. That is, for the parameter vector
\[ 
\btheta = \left(\theta_{1}, \hdots, \theta_{d}, \hdots, \theta_{D}\right)
\]
the dimensionality of the parameter $\btheta$ cannot exceed the number of data points $K$ (i.e.  $D \leqslant K$). An example of such a joint model is the matrix normal model given by
\[  
\mathrm{vec}(\mathbf{Y}) \sim \mathcal{N}_{NP}(\mathrm{vec}(\mathbf{M}), \Sigma \otimes \Omega)
\]
where $\mathbf{M}$ is $N \times P$; $\Sigma$ is $N \times N$; $\Omega$ is $P \times P$; and $\otimes$ denotes the Kronecker product. Accordingly, our parameter vector of interest would consist of
\[ 
\btheta= [\mathrm{vec}(\mathbf{M)}, \mathrm{vech}(\Sigma), \mathrm{vech}(\Omega)]
\]
with a dimension equal to $\frac{2NP+N(N + 1) + P(P + 1)}{2}$ - which is always bigger than $N \times P$. In general, note that when $ P \gg N$, the number of features is greater than the number of observations, the measurement density $\alpha = \left( \frac{N}{P} \right)$ is finite. When $\alpha $ does not converge to infinity as $N \to \infty$, we are faced with a high-dimensional inference problem for which much of theory of classical statistics does not apply \citep{Advani2016}.

Now consider the opposite scenario. If we have no grouping information about the $K$ measurements then the most reasonable thing to do is to assume exchangeablility and thus treat all the $K$ measurements as if drawn from a common distribution $p_{\btheta}$
\[ 
 y_k \overset{i.i.d.}{\sim} \quad p_{\btheta}  
\]
Compared to the previous situation in which our data consisted of a single $K$-dimensional vector observation, here we have a total of $K$ one-dimensional scalar observations. With so many samples, inference should be much more manageable. Furthermore, we now have much more flexibility in choosing the dimension of $\btheta$. The dimensionality of the parameter vector $\btheta$ controls the flexibility of the model's distribution. For instance, we can specify a low information capacity model such as a univariate Gaussian model with unknown mean. In this case, $\btheta$  would be one dimensional. At the other end of the spectrum, we can specify a very flexible kernel density estimator to learn the projected data's density. Regardless, by collapsing high-dimensional data into a one-dimension marginal distribution, we would still be unable to learn the global patterns of dependencies that govern the data distribution in its original high dimensional space.

In most statistical studies however, we would expect the data matrix to be characterized by a grouping structure that preserves the distinction between a feature and an observation, such that the rows of the matrix correspond to the observations and the columns correspond to the $P$ features. The  grouping structure implicit in a data matrix effectively restricts how the indices can be shuffled and transformed. Nonetheless, the data matrix can still be characterized by complicated grouping structures. For example, the  $P$ variables can be partitioned into a set of views (i.e. data sets) after which the observations within a view can be partitioned into a set of clusters or sub-populations \citep{mansinghka2016}.

In what follows, we will treat the $N$ rows of the matrix as observations and the $P$ columns as features such that the joint distribution is given by
\[ 
p\left(\by_1, \by_2, \hdots, \by_N, \btheta\right) 
\]
We can also partition the $P$-dimensional observation vector into $G$ multiple groups, where $ G \leq P$.
\[ 
\by_n= \left(\by_{n,1}^\intercal, \hdots,\by_{n,g}^\intercal , \hdots, \by_{n,G}^\intercal\right)
\]
Or even partition the $N$ observations into $C$ clusters such that 
\[ 
\by = \big(\{\by_n\}_{n \in N_1}, \hdots , \{\by_n\}_{n \in N_C}\big)
\]
To obtain more complicated grouping structures, we can cluster the observations after we group the features such that observations for group $g$ would  belong to  cluster $C_g$. The data can then be grouped as such. 
\[ 
\by = \{\by_{(n,g)}\}_{n \in N_{C_g}}
\]

By partitioning the data matrix into groups and assuming partial exchangeability, we can build a hierarchical model for all the groups. In a hierarchical model, the measurements in each group are exchangeable within a corresponding sub-model and the parameters of the sub-models are exchangeable in their prior distribution \citep{gelman2014}. 

\subsection{What is a Latent Variable?}

Many definitions for a latent variable have been proposed in the literature \citep[see][for a review]{Bollen2002}. Although the definitions that have been put forward range from the informal to the mathematical, many of them rely on distinctions that can be conceptually problematic when working in a probabilistic modeling framework. For example, defining a variable as \textit{latent} based on whether it is treated as random rather than fixed does not carry over to the probabilistic framework where all variables, known or unknown, are treated as random. Furthermore, definitions that are based on the notion of unobservability introduce some conceptual issues that complicate any attempt to give a simple definition based on that concept alone. In  a probabilistic model, a variable is either observable or unobservable, but not all unobservable variables in a probabilistic model correspond to what we think of as a latent variable when we are working outside the probabilistic framework. In particular, definitions based on unobservability  break down when considering how to treat the random noise term $\bepsilon_n$, whether it is found in a simple regression model or in an implicit model that transforms the distribution of the error into a more complicated  distribution $g(\bepsilon_n;\btheta) $ \citep{goodfellow2014}. In both cases, the error term is an unobservable random quantity whose distribution deterministically induces a distribution on the response $\by_n$ through an injective function $g$ whose inverse is $g^{-1}$. Nonetheless, it cannot be considered a latent variable since the posterior
\begin{align} 
p(\bepsilon_n \vert \by_n)= \mathbb{I}[\bepsilon_n = g^{-1}(\by_n)]
\end{align}
is a point mass, a deterministic not a random function \citep{tran2017}. Note that for a function to have an inverse, the dimension of $\bepsilon_n$ should equal the dimension of $\by_n$. It could be argued that for variable to be considered a latent variable it needs be an unobserved random variable whose posterior distribution is not degenerate.  For example, in the Tobit model \citep{tobin1958}
\begin{align}
    \begin{split}
\epsilon_i \sim N(0 ,\sigma^2)\\
y_i^* = \beta x_i +\epsilon_i\\
y_i = max(0, y_i^*)
    \end{split}
\end{align}
the error term $\epsilon_i$ accordingly would not be considered a latent variable since the function $g(x)=m+sx$ is invertible given that both $\epsilon_i$ and $y_i^*$ are scalars (i.e. both of dimension one). In contrast, $y_i^*$ can be considered a latent variable since its posterior is not a point mass and the function $g(x) = max(0, x)$ is not invertible, even though both $y_i^*$ and the observed variable $y_i$ are scalars. To complicate things further, one could argue that in the Tobit model, we are in fact dealing with a partially observable  variable model $y_i = max(0, \beta x_i +\epsilon_i)$ rather than with a latent variable model per se.

Requiring a latent variable to be an unobservable random variable \textit{with a non-degenerate posterior distribution} can provide us with an adequate working definition of a latent variable. Still we argue that a latent variable can be better understood by the function it serves in a model. In particular, we propose that the inclusion of a latent variable in a statistical model allows us to accomplish two tasks: First, to capture statistical dependencies in the data and second, to learn a latent structure that potentially governs the data generating mechanism for the observations (i.e. learn the posterior distribution $p(\bz \vert \by)$). With this in mind, note that the notion of a latent variable as widely understood in the majority of modeling context is most intuitively explained  by the \textit{local independence definition} of a latent variable \citep{lord1952}. We qualify the preceding statement by the word \textit{majority} because in the local independence definition does not always hold for non-probabilistic models such as SEM where the latent variables can regressed on each other to induce dependencies among the latent variable. Furthermore, the local independence definition might not be as intuitive in the case of models, whether estimable or not, in which there are a great number of latent variables that combine together to give rise to an observed response  of a much lower dimension \citep{Thomson1916}. 

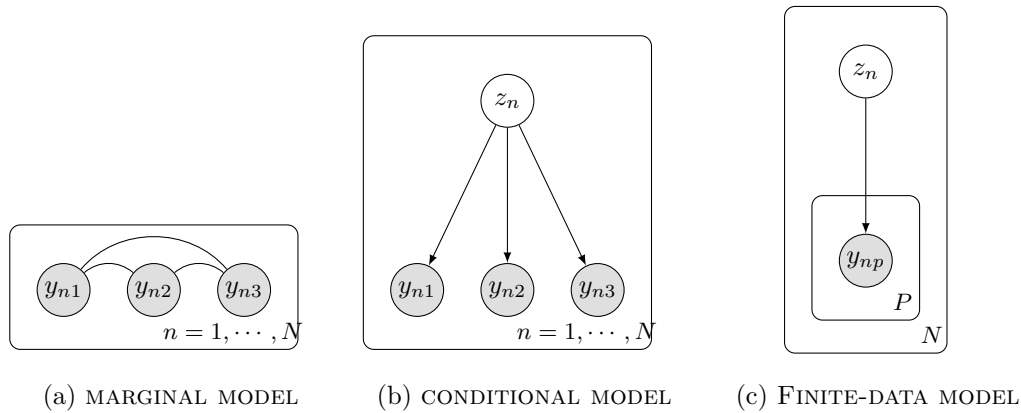
\begin{figure}
\begin{subfigure}[t]{0.3\columnwidth}
  \centering
  \begin{tikzpicture}[x=1.7cm,y=1.8cm]

  \node[obs , xshift=-1.2cm]                                (y1)      {$y_{n1}$} ;
  \node[obs]                              (y2)      {$y_{n2}$} ;
  \node[obs, xshift=1.2cm]                              (y3)      {$y_{n3}$} ;

    \path 
        (y1) edge [bend left=40] node [] {} (y2)
        (y1) edge [bend left=50] node [] {} (y3)
        (y2) edge [bend left=40] node [] {} (y3);

  \plate[inner sep=0.35cm, yshift=0.15cm,
  label={[xshift=-55pt,yshift=14pt]south east: $n=1,\cdots,N$}] {plate2} {
    (y1)(y2)(y3)
  } {};

\end{tikzpicture}
  \vspace{1ex}
  \caption{\textsc{marginal model}}
  \label{sub:dependent}
\end{subfigure}
\begin{subfigure}[t]{0.3\columnwidth}
  \centering
  \begin{tikzpicture}[x=1.7cm,y=1.8cm]

  \node[latent]                                         (z)      {$z_n$} ;
  \node[obs, below=of z , xshift=-1.2cm]                                (y1)      {$y_{n1}$} ;
  \node[obs, below=of z]                              (y2)      {$y_{n2}$} ;
  \node[obs, below=of z, xshift=1.2cm]                              (y3)      {$y_{n3}$} ;

  \edge{z}{y1, y2, y3};

  \plate[inner sep=0.35cm, yshift=0.15cm,label={[xshift=-55pt,yshift=14pt]south east: $n=1,\cdots,N$}] {plate2} {
    (z)(y1)(y2)(y3)
  } {};

\end{tikzpicture}
  \vspace{1ex}
  \caption{\textsc{conditional model}}
  \label{sub:latent_pop}
\end{subfigure}
\begin{subfigure}[t]{0.3\columnwidth}
  \centering
  \begin{tikzpicture}[x=1.7cm,y=1.8cm]

  \node[latent]                                         (z)      {$z_n$} ;
  \node[obs, below=of z]                                (y)      {$y_{np}$} ;

  \edge{z}{y};

  \plate[inner sep=0.35cm, yshift=0.15cm,
    label={[xshift=-14pt,yshift=14pt]south east:$P$}] {plate1} {
    (y)
  } {};
  \plate[inner sep=0.35cm, yshift=0.15cm,
    label={[xshift=-14pt,yshift=14pt]south east:$N$}] {plate2} {
    (z)(y)(plate1)
  } {};

\end{tikzpicture}
  \vspace{1ex}
  \caption{\textsc{Finite-data model}}
  \label{sub:latent_samp}
\end{subfigure}
\caption{\textbf{(a)} Marginal graphical model for the three dependent variables. \textbf{(b)} Directed graph of the observed variables conditional on the latent variable. \textbf{(c)} The latent variable model as a directed graphical model for a sample of $N$ multivariate observations, each a $P$-dimensional vector.  
}
\label{fig:vgp}
\end{figure}

To illustrate the basic idea, consider the following toy example. We observe $N$ \textit{multivariate observations}, $N$ samples of three dependent variables $\lbrace(y_{n1}, y_{n2}, y_{n3})\rbrace_{n=1}^N$ (Figure \ref{sub:dependent}). The inclusion of a scalar latent variable $z_n$ for the $n$th observation allows us to factorize the joint distribution of ($y_{n1}, y_{n2}, y_{n3}$) into a set of conditionally independent factors, as shown by the directed graph in Figure \ref{sub:latent_pop}. For example, if we have a set of $P$ dependent variables where the dependency is restricted to positive linear association among them, then conditioning on just one single latent variable can be sufficient to explain the pattern of dependencies in the $P$ dimensional data. The resulting latent variable model for the  $N$ observations can be concisely expressed by the directed graph in Figure \ref{sub:latent_samp}.

\subsection{What is a Probabilistic Latent Variable Model?} 

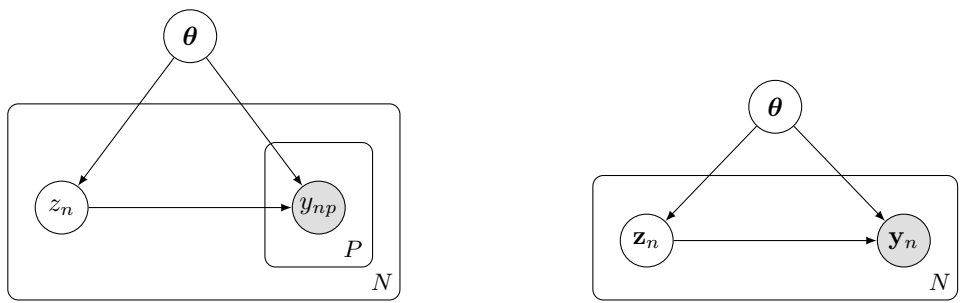
\begin{figure}[htb]
\begin{subfigure}[t]{0.49\columnwidth}
\centering
\begin{tikzpicture}[x=1.7cm,y=1.8cm]

  
    \node[latent]                                     (theta)   {$\btheta$} ;
  \node[latent, below left=of theta, yshift=-.5cm]                 (z)      {$z_n$} ;
  \node[obs, right=of z, below right=of theta, yshift=-.5cm]       (y)       {$y_{np}$} ;

  \edge{z}{y};
  \edge{theta}{y};
  \edge{theta}{z};

  \plate[inner sep=0.35cm, yshift=0.15cm,
    label={[xshift=-15pt,yshift=14pt]south east:$P$}] {plate1} {
    (y)
  } {};
  \plate[inner sep=0.35cm, yshift=0.15cm,
    label={[xshift=-15pt,yshift=14pt]south east: $N$}] {plate2} {
    (z)(y)(plate1)
  } {};

\end{tikzpicture}
\vspace{1ex}
\caption{Unidimensional Latent Variable Model}
\end{subfigure}
\begin{subfigure}[t]{0.49\columnwidth}
\centering
\begin{tikzpicture}[x=1.7cm,y=1.8cm]

  \node[latent]                                     (theta)   {$\btheta$} ;
  \node[latent, below left=of theta]                 (z)      {$\bz_n$} ;
  \node[obs, right=of z, below right=of theta]       (y)      {$\by_{n}$} ;

  \edge{z}{y};
  \edge{theta}{y};
  \edge{theta}{z};

  \plate[inner sep=0.35cm, yshift=0.15cm,
    label={[xshift=-15pt,yshift=14pt]south east: $N$}] {plate} {
    (z)(y)
  } {};

\end{tikzpicture}
\vspace{1ex}
\caption{Multi-dimensional Latent Variable Model}
\end{subfigure}
\caption{\textbf{(a)} For each sample observation $n$, $z_{n}$ is a local latent variable, and the $y_{np^{\prime}}$ are $P$ observed variables. $\btheta$ is a vector of global latent variables (parameters) \textbf{(b)} Directed graph of the observed variables conditional on the latent variable.  
}
\label{fig:lvm}
\end{figure}

Let us define \textit{probabilistic latent variable model} as a statistical model of observable variables that incorporates unobservable variables. Accordingly, any parametric or nonparametric hierarchical Bayesian models subsumed under the decomposition  given by Equations \ref{eq:parm_bayesian_model} and \ref{eq:nonparm_bayesian_model} can be considered as special cases in the general class of probabilistic latent variable models. To formulate a general modeling framework for the family of probabilistic latent variable models, we incorporate a \textit{D}-dimensional \textit{local latent} variable vector $\bz_n=(z_1, \hdots,z_D)$ for each observation $\bz_n$ and a \textit{M}-dimensional \textit{global latent} variable vector $\btheta =(\theta_1, \hdots,\theta_M)$ shared across all observations \citep{blei2016}. The local latent variable, $\bz_n$, serves the function of capturing local dependencies \textit{within} the observation vector $\by_n$ (e.g. clusters or common factors) whereas the global latent variable $\btheta$ captures the dependencies \textit{across} all, or even a subset, of the $N$ observable random vectors. Accordingly, the joint probability distribution for a probabilistic latent variable model can be expressed as such
\begin{align}\label{eq:joint}
p(\by_1,\cdots, \by_N, \bz_1,\cdots, \bz_N, \btheta) 
\end{align}
Now, there are countless many ways we could factor the joint distribution given by Equation \ref{eq:joint}. Any chosen factorization encodes a set of implicit assumptions that make the particular decomposition possible. For example, if the observations in $\by = \lbrace \by_1, \by_2, \cdots, \by_N \rbrace$ are sequential in nature such that their distribution is no longer invariant under permutation of indices (i.e. exchangeable), but is still invariant under shifts (i.e. stationary), then a very general class of temporal generative models \citep{gemici2017} can be formulated by the following decomposition. 
\begin{align}\label{eq:GTM}
p(\by_{\ \leq N},\bz_{\ \leq N}, \btheta) = p(\btheta)\prod_{n=1}^N\ p(\by_n \mid f_x(\bz_{\ \leq n},\by_{\ < n}), \btheta)p\big(\bz_n \mid f_z(\bz_{\ <n},\by_{\ <n}), \btheta\big)
\end{align}
The family of generative temporal models give by Equation \ref{eq:GTM} include hidden Markov models, non-linear state space models \citep{tornio2007}, and the Variational Recurrent Neural Network (VRNN) \citep{chung2015}. If we are not modeling temporal or sequential data, then we can assume that the \textit{N} multivariate observations  are  indeed \textit{exchangeable} and obtain the following decomposition instead.
\begin{align}\label{eq:latent_general}
p(\by_1,\cdots, \by_N, \bz_1,\cdots, \bz_N, \btheta) = p(\btheta)\prod_{n=1}^N\ p(\by_n \mid \bz_n,  \btheta)p(\bz_n\mid \btheta)
\end{align}
This  decomposition subsumes a very broad class of latent variable models such as hierarchical Bayesian regression models \citep{gelman2014}, Bayesian mixture models, additive matrix decomposition models (e.g. factor analysis, PCA, and CCA), and Bayesian nonparametric models. A graphical representation of this general latent variable family of models is shown in Figure \ref{fig:lvm}.

Notice that the decomposition given by Equation \ref{eq:latent_general} implies a second assumption; specifically, that the $N$ local latent variables $\bz_n$ are exchangeable in their joint distribution. The joint distribution here refers to the prior $p(\bz_n\mid \btheta)$ which is governed by the global latent variable $\btheta$. The role of global latent variable in the model is to control the extent of information pooling \emph{across} the observations. To illustrate, if we set the variance of the prior distribution to infinity, we obtain a separate model for each group as a result. That is, a different model is fit to each observation vector, where $\by_n$ can be thought of as an ``experiment'' with $P$ measurements. On the other extreme, setting the variance to zero would result in \textit{complete pooling} of information, effectively giving us a single model for all $N \times P$ collapsed observations \citep{gelman2014}. Incidentally, in the context of mixed effects models, it is interesting to point out that a fixed effect can be thought of as a special case of a random effect, but one with an infinite variance prior.

\subsection{Implicit Vs Explicit Models} Most of the time, we propose models whose likelihoods and priors have an explicit closed-form probability densities. In some occasions, however, if we are able to sample from a simulator that is better at modeling the generative mechanism that gives rise to the data, then we can refrain from specifying an explicit probability density for either the data distribution or the prior. Models that do not specify explicit densities are known in the literature as \textit{implicit models}, first introduced by \citet{diggle1984} to approximate an intractable likelihood and later extended to approximating the posterior distribution by \citet{tavare1997} in a method that is now known as Approximate Bayesian Computation (ABC) \cite{pritchard1999}. In an implicit model, instead of defining a likelihood, we define a deterministic function $f(.)$ whose input is a random noise source $\bepsilon_n$ and whose output is the observation
\begin{align}
\by_n= f(\bepsilon_n \mid \bz_n,  \btheta); \quad \bepsilon_n \sim \ s(·)
\end{align}
The function $f(.)$ can for example be implemented as a deep neural network. 
Furthermore, the density for the local unobserved variable can also be defined implicitly as such  
\begin{align}
\bz_n= f(\bepsilon_n \mid \btheta); \quad \bepsilon_n \sim \ s(·)
\end{align}
 We can sample from the posterior $q(\mathbf{z}_n \mid \mathbf{y}_n;\bnu)$ for observation $n$
\begin{align} 
\bz_n \sim  \mathcal{N}_d\big(\bz_n \vert \mathbf{NN}_{\mu}(\by_n;\bnu), \mathbf{NN}_{\sigma}(\mathbf{y}_n;\bnu)\big)
\end{align}
by using the implicit function $g$
\begin{align}
\bz_n = g(\epsilon_n,\by_n; \bnu) = \mathbf{NN}_{\mu}(\by_n;\bnu) + \mathbf{NN}_{\sigma}(\mathbf{y}_n;\bnu) \odot \epsilon_n
\end{align}
where $\odot$ denotes the elementwise product and
\begin{align}
\epsilon_n \sim \mathcal{N}_d\big(0 ,\bI\big)
\end{align}
If the neural networks $\mathbf{NN}$ are not invertible, as is usually the case, then the function $g$ is not invertible and the likelihood is therefore intractable (i.e. implicit). Almost all latent variable models can be considered implicit models in the sense that their marginal distributions do not have an explicit form. The most notable exception would be a factor analysis model for which we have an explicit form for the marginal distribution after the latent variables get integrated out. 

\subsection{Deep Vs Shallow Models} Also note that the decomposition given by Equation \ref{eq:latent_general} above presumes that the elements of the local latent variable vector $\bz_n$ form a single layer. \emph{Deep models} in contrast contain several hidden layers that form a rich hierarchical structure. To obtain a joint factorization that includes deep models as a general case, we just need to divide the vector of local unobserved variables $\bz_n$ into $L$ subsets $\bz_n=\lbrace \bz_{n,1}, \,\cdots, \bz_{n,L} \rbrace$, such that each  $\bz_{n,1}$ corresponds to a separate level in the sampling hierarchy. Unlike shallow models, deep models with a hierarchy of latent variable dependencies are better able to capture rich latent structure \citep{ranganath2015}.

\section{Brief History of Latent Variable Models}

Although latent variable modeling is a branch of multivariate statistics, many of the field's major developments originated and were motivated by research problems, not within statistics, but rather from the psychological and social sciences \citep{izenman2008}. Whereas the early breakthroughs mainly came from psychology, educational measurement, and economics, the more recent advances are increasingly coming from the field of machine learning, especially from the very active research area of unsupervised learning in which researchers are trying to accomplish the goal of endowing computers the ability to learn complex data patterns in an unsupervised automatic manner \citep{lake2015,rezende2016}. 

The quest to understand learning and intelligence, whether in humans or machines, has always been a motivating force in the field from its early beginnings. Indeed, the psychologist Charles Spearman, influenced by Francis Galton's development of the correlation coefficient, introduced \emph{factor analysis} in 1904 as a statistical method to study the organization of mental abilities in humans \citep{Spearman1904}. He proposed that from the correlations of multiple observed variables (e.g. tests of general ability), we can infer a single latent factor that can explain the common variance shared among all the observations. He referred to the latent variable common to all the measured ability tests as \emph{g}, the general factor of intelligence. 

\subsection{Summary of Key Developments}

The factor analytic approach paved the way for, what seemed to many in the social and psychological sciences, an objective and quantitative framework for measuring unobservable \emph{ hypothetical constructs} such as ``self-esteem'', ``verbal ability'', and ``quality of life''. Indeed Spearman's single factor model, in which both the one latent variable and the observed variables are assumed continuous, has laid the foundation for a long research program in psychometrics that has spanned well over a century. To account for the inadequacy of a single factor to account for the covariance pattern in testing data, the psychologist Thurstone extended the single factor model to multiple factors \citep{Thurstone1947}. The factor model was constantly refined and generalized, culminating in the \emph{structural equation model} introduced by \citep{Joreskog1970} in which linear structural relationships between the latent variables are also modeled. In educational testing, \emph{item response theory} models \citep{Birnbaum1968} were developed to accommodate categorical observed variables encountered in testing in which test items assume two values, correct and incorrect. In sociology, clustering analysis motivated the development of \emph{latent class models} (i.e. mixture models) in which the latent variables are assumed to be categorical \citep{lazarsfeld1950}. Since then, a large body of work has been built on top of these foundational early advances and many authors \citep{muthen2002, bartholomew2011, hwang2014} have attempted to bring together the various models under one roof in a general framework. The most encompassing as of yet is the Generalized Linear Latent and Mixed model (GLLMM) introduced by  \citet{rabe2004} as a unified general framework that brings together not just factor analysis, IRT, structural equations, and latent class models commonly encountered in psychometrics and social quantitative methods, but also mainstream statistical random effects models such as multilevel regression, and longitudinal models. 

\subsection{The Statistical Myth of Measuring Hypothetical Constructs}

The enthusiasm regarding the factor analytic approach to uncover underlying theoretical constructs was not universally shared. In particular, the interpretation of a latent variable as a hypothetical construct was much criticized. One of the earliest criticism came from Spearman's colleague, Karl Pearson, widely regarded as the founder of modern mathematical statistics. The two men constantly feuded with each other throughout their careers due to their opposing scientific philosophies \citep{Hagglund2001}. Whereas Spearman strove to discover the fundamental laws of psychology and invent a method by which he can objectively measure latent traits such as intelligence, Pearson did not see a place for unobserved and subsequently unmeasurable phenomena in the scientific enterprise. Incidentally, \citet{Pearson1901} had pioneered a method closely related to, and often conflated with Factor Analysis (FA) - \emph{principle component analysis} (PCA). Pearson's PCA, which was later formulated as multivariate technique by \citet{Hotelling1933}, was introduced not as an explanatory model, but as a dimensionality reduction technique in which the data is geometrically transformed (i.e. projected) into a lower dimensional subspace.

The criticisms nonetheless did not end with Pearson. The English educational psychologist \citet{Thomson1916} proposed a sampling model that generated the same correlation structure as a general factor model but that did not assume a single factor. In particular, whereas Spearman's model, and later extensions, proposed that positive correlations among \textit{P} variables suggest the existence of one or more factors fewer than \textit{P}, Thomson's sampling model assumed the opposite: that there are more factors than we have variables and that each variable is the result of a great number of independent causes, much greater than \textit{P}. Even though it seems more plausible to assume that there are a great number of factors (mental subprocesses) that contribute to a score on any one ability test or item questionnaire, psychometricians following Spearman's footsteps ignored these criticisms and pursued a research program based on the indefensible premise that factor analysis and its extensions are statistical methods that can explain correlations in the observed  data, rather than just describe them. It can be argued that the explanatory approach in which the existence of a hypothetical construct is inferred by performing exploratory analysis on limited sample self-report data cannot substitute the rigorous process of developing operational definitions of unobservable constructs on theoretical grounds or even on experimental measurements of the cognitive processes underlying the observed responses. Indeed, throughout the history of psychometrics, there have been repeated calls to guard against the \emph{reification fallacy} in which a latent variable (e.g. mental ability) that captures a common variability in the data is treated as if it were a physical quantity (e.g. height) that is amenable to direct measurement \citep{gould1996,sobel1994}. 

Moreover, given that the definition of a statistical model provided in section \ref{sect:statmodel} does not say much about the nature of the relationship between a statistical model and its target phenomenon observed in the physical universe, we would like to point out here that the modeling perspective we adopt is founded on the premise that a statistical model is a \emph{phenomenological model} of reality describing relationships among observed variables in a mathematical formalism that is not derived from first principles \citep{hilborn1997}. We consider a model phenomenological in the Kantian sense \citep{kant1998} in which the word \emph{phenomenon} denotes an empirical object that is known through the senses and whose mental representation is constructed by experience. The \emph{phenomenon} is contrasted with the \emph{noumenon}, the ``thing-in-itself'' that is not directly knowable by the senses but which can be known by the mind in the form of a theoretical construct. For example, whereas the electromagnetic force is a theoretical construct arrived at from first principles and provides a causal explanation to empirical observations (e.g. effects of magnetic and electric fields), the empirical observations themselves constitute experienced phenomenon whose correlative nature can provide nothing more than evidence for noumenal knowledge.

\subsection{The First Latent Variable Model}

If we restrict latent variable models to those models in which latent variables denote hypothetical constructs, arguably Spearman's factor model could be considered the first formulation of a latent variable model. However, such a restriction based on a narrowly defined interpretation is not universally accepted neither within psychometrics, the field Spearmen helped start, nor within the larger statistics community \citep{Skrondal2004}. If we just consider the statistical form of the model, then  it could be argued that the first latent variable model should be attributed to the astronomer Airy, who in 1861  proposed a variance components model (i.e. random effects model) for the repeated measurement of an astronomical object over several nights \citep{airy1861}. Both Airy's variance components model and Spearman's factor model have the same graphical representation depicted in Figure \ref{sub:latent_samp}. The main difference lies in that the latent variable in Airy's model is a random effect that captures unobserved heterogeneity whereas in Spearman's model the latent variable denotes a hypothetical construct which we seek to measure. 

In his model, Airy makes $P$ repeated measurements of an astronomical phenomenon for each of $N$ nights. See \citet{searle2009} for a more details and the original formulation. Assuming a balanced design, the model can be expressed as a probabilistic generative model by the following two-stage sampling hierarchy.
 
\begin{align}
    \begin{split}
\by_n \mid  z_n &\sim\ \mathcal{N}_P(\bone\mu + \bone z_n,\, \sigma^2_{\epsilon}\mathbf{I}) \\
z_n &\sim \mathcal{N}(0, \sigma^2_{z})  \quad \textit{for } n=1, \cdots,N 
    \end{split}
\end{align}
where $\by_n \in \mathbb{R}^{P}$ is the vector of repeated measurements and $z_n \in \mathbb{R}$ is a latent variable (i.e. random effect).

\subsection{Latent Variable Models in Machine Learning}

Although the GLLMM framework does indeed treat random coefficient models (which include hierarchical and longitudinal models) as latent variable models, the framework leaves out important multivariate data reduction methods such PCA, Independent Component Analysis (ICA), Canonical Correlation Analysis (CCA) from the formulation. The argument put forward by \citet{Skrondal2004} contends that PCA and similar procedures are not latent variable models because they cannot be formulated as statistical models with latent variables corresponding to lower dimensional space. However, their assertion does not seem to have taken into consideration results that had been previously published in the literature. More specifically, in recent years, research in latent variable modeling has undergone a very active resurgence in the field of unsupervised machine learning. Indeed, great progress has been made in developing new latent variable models and recasting existing clustering and dimensionality reduction methods in a probabilistic graphical model formulation. For example, building on previous research in neural networks \citep{mackay1995a,bishop1998}, \citet{tipping1999} demonstrated that PCA can be indeed be formulated as a probabilistic latent variable model. In their paper, they showed that Probabilistic PCA is a special case of Factor Analysis (FA). More specifically, whereas in FA the error covariance matrix is diagonal, in Probabilistic PCA, it is $\sigma\bI$ and standard PCA is obtained when $\sigma \rightarrow 0$. \citet{Roweis1999} review several common methods showing that not just PCA and FA, but also ICA, Kalman Filters, and Hidden Markov Models (HMMs) can be considered as special cases of a general linear-Gaussian latent variable model. A probabilistic formulation of CCA \citep{bach2005} later followed. Recently, a Group Factor Analysis (GFA) model has been proposed by \citet{klami2015} that generalizes Probabilistic CCA to more than two data sets.  

It is these developments that motivate the material in the next two sections. In the social sciences, latent variable models have been traditionally been viewed and used as tools to study hypothetical constructs. However, ongoing research in machine learning, deep learning, and probabilistic graphical modeling has given a new life and meaning to latent variable modeling. These recent developments present latent variables as capable of endowing statistical models not only with the ability to perform essentials data analysis tasks such as dimensionality reduction, statistical clustering, and accounting for unobserved heterogeneity, but also with explanatory and interpretative powers. The inclusion of latent variables allows us to capture statistical dependencies among observed variables and thus learn the hidden structure underlying the data generating mechanism. This is especially true when latent variable models are cast in a generative probabilistic framework, a perspective that the next section attempts to describe in detail. The following sections forms the foundation for subsequent sections where many of the most commonly encountered latent variable models are constructed from the most basic elements.

\section{Constructing Probabilistic Latent Variable Models}\label{sec:lvm}

In this section, we will attempt to show how starting from a joint probability density, we can formulate a spatiotemporal models, multivariate models, regression models, and latent variable models.  We will also attempt to describe how regression models are related to latent variable models through reduced rank regression. We end the section by giving an overview of latent variable models most commonly encountered in practice. 

To begin, let us consider a two-group partition of the observation vectors that populate the data matrix $\mathbf{Y}$.  To simply notation, we can effectively create a two-group partition by concatenating the data matrix  with a matrix of co-variates such that each row would consist of a \textit{P}-dimensional observation vector $\by_n=(y_{n1}, \cdots, y_{nP})$ and an \textit{M}-dimensional covariate vector $\bx_n=(x_{n1}, \cdots, x_{nM})$. The concatenated data would be an $N$ by $(P+M)$ matrix consisting of a total of $ N\times(P+M)$ measurements 
\begin{align}
\begin{bmatrix}
(\by_1^{\rm T}, \bx_1^{\rm T} ) \\
\vdots \\
(\by_n^{\rm T}, \bx_n^{\rm T} )\\
\vdots \\
(\by_N^{\rm T}, \bx_N^{\rm T})
\end{bmatrix} 
=\begin{bmatrix}
    y_{11}  & \dots  & y_{1P} & x_{11}  & \dots  & x_{1M}\\
    \vdots  & \ddots & \vdots & \vdots  & \ddots & \vdots \\
    y_{i1}  & \dots  & y_{nP} & x_{n1}  & \dots  & x_{nM} \\
    \vdots  & \ddots & \vdots & \vdots  & \ddots & \vdots  \\
    y_{N1}  & \dots  & y_{NP} & x_{N1}  & \dots  & x_{NM}
\end{bmatrix} 
\end{align}
with a joint distribution given by
\[
p\big(\left(\by_1,\bx_1\right), \cdots,\left(\by_N, \bx_N\right), \btheta\big)
\]

\subsection{Spatiotemporal Models}  

We can treat the data as spatiotemporal in nature by letting the row index denote time and mapping the column index to a given spatial coordinate. Using a spatio-temporal mode, we can learn both the temporal and spatial dependencies of the data. In a spatiotemporal model we do not treat the observations that populate the data matrix as exchangeable neither in the temporal nor in the spatial dimension. In the spatial dimension, the patterns of dependencies tend to vary in complexity depending on the type of data. In the temporal dimension in contrast, the observations tend to be  characterized by a less complex pattern of sequential dependencies. For example, the observation $\by_n$ can be dependent on the preceding observation $\by_{n-1}$ only.

If we wanted a very general model, we can formulate a non-separable spatio-temporal model in which the spatial and temporal components are allowed to interact. If we assume some form of stationarity in the temporal process such that the joint probability distribution of a finite sequence of consecutive observations does not change when they shifted in time, then a family of non-separable spatio-temporal models can be specified by the following decomposition.
\begin{align}\label{eq:nonsep}
p(\by_{\ \leq N};  \btheta) =\prod_{n=1}^N\ p(\by_n \mid f(\by_{\ < n}); \btheta)
\end{align} 
where we let the subscript $n$ index time such that the response $\by_n$ can be dependent on  preceding responses $\by_{\ < n}$ through a nonlinear function $f$.

Such a model is commonly encountered and is appropriate for modeling dynamical systems in which the current spatial component of the process evolves as a function of past spatial locations. Although non-separable models tend to be realistic for many real world applications, their estimation can be computationally intractable, especially in the case when the dimensionality of \textit{N} or \textit{P} is high.

Alternatively, when adequate motivation exists, a separable spatio-temporal model can be used instead. We can formulate a separable model by assuming that the parameters that govern the distribution of the temporal component are independent of those that govern the spatial component. To obtain a separable model, we let $\btheta = (\btheta^{S}, \btheta^T)$ such that $\btheta^{S}$ governs the distribution of the $P$ elements of $\by_n$ whereas $\btheta^T$ governs the distribution of the $N$ observations. 
\begin{align}\label{eq:sep}
p(\by_{\ \leq N}; \ \btheta^{S}, \btheta^T) =\prod_{n=1}^N\ p(\by_n \mid f_{\btheta^T}(\by_{\ < n}), \ \btheta^{S})
\end{align} 

\subsection{Multivariate Models} 

A spatiotemporal model can be viewed either as a temporally varying spatial model or as spatially varying time series model \citep{banerjee2014}. In either case, the data modeled is assumed to be characterized by both spatial and temporal dependencies. In contrast, in multivariate models, we relax the stationarity assumption that allows us to model the sequential dependencies in the temporal dimension. Instead,  we assume that the observations are exchangeable in their joint distribution. Exchangability  give us the following factorization.
\[
p\big(\left(\by_1,\bx_1\right), \cdots,\left(\by_N, \bx_N\right), \btheta\big)= \prod_{n=1}^N p\big(\left(\by_n,\bx_n\right)\mid \btheta \big)p(\btheta)
\]

Note that the parameter vector $\btheta$ governs the distribution of the pairs of $\by_n$ and $\bx_n$ jointly. However if we wish to predict $\by_n$ from $\bx_n$, we need to assume that the parameters that govern the distribution of $\by_n$ are independent of those that govern the distribution of $\bx_n$. That is, given the partition $\btheta = (\btheta^{Y}, \btheta^{X})$, we have
\[
p(\btheta^{Y}, \btheta^{X})=p(\btheta^{Y}) p(\btheta^X)
\]
and we can factorize the joint distribution as such.
\begin{align}
p\big(\left(\by_1,\bx_1\right), \cdots,\left(\by_N, \bx_N\right), \btheta\big)=\prod_{n=1}^N p(\by_n, \mid f_{\theta}(\bx_n), \btheta^{Y})\prod_{n=1}^N p( \bx_n \mid \btheta^{X}) p(\btheta^{Y}) p(\btheta^X) 
\end{align}
The factorization implies that the distribution of $\bx_n$ does not influence the conditional distribution of $\by_n$. The assumption allows us to ignore the distribution of $\bx_n$ when the focus is on determining the parameter $\theta^Y$ - such as the case in many regression or path analysis models for example, where $\bx_n$  is considered an exogenous variable \citep{wright1921}. 

\subsubsection{Normal Linear Models} 

Let us split the vector of parameters into two components such that 
$$\theta^{Y} = [vec(\bB),\  vech(\bSigma)]$$
where 
\[
\bB=\begin{bmatrix}
\bbeta_1^{\rm T} \\
\bbeta_2^{\rm T} \\
\vdots \\
\bbeta_P^{\rm T}
\end{bmatrix} =
\begin{bmatrix} \beta_{11} & \cdots & \beta_{1m} \\
\beta_{21} & \cdots & \beta_{2m} \\
\vdots & \ddots & \vdots \\
\beta_{p1} & \cdots & \beta_{pm}
\end{bmatrix}
\]
and $\bSigma$ is positive definite matrix, and  let us further assume that 
\begin{align*}\mu(\bx_n; \bB)= \bB \bx_n \quad \textit{and}  \quad\bsigma(\bSigma)= \bSigma
\end{align*}
we arrive at following data distribution for a linear regression model
\begin{align*}
p\left(\by_n \mid \bx_n, \theta^{Y}\right) \equiv \sN_p\left(\by_{n} \mid
\bB\bx_n , \  \bSigma\right)
\end{align*}
Note that since both $\by_n$ and $\bx_n$  are multivariate vectors, $\mu(\bx_n; \bB)$ is vector-valued function that transforms $\bx_n$ into $\by_n$. Also note that we dropped the index $n$ from both functions.  The joint probability distribution can now be factored as such. 
\begin{align}\label{multivariate_normal}
 \prod_{n=1}^N  \sN_p( \by_{n} \mid \bB \bx_n, \ \bSigma) p\left(\bB, vech(\bSigma)\right)\prod_{n=1}^N\ p(\bx_n\mid \theta^{X})p(\theta^{X})
 \end{align} 
 
Let us simplify things further and let $\bSigma=\Diag(\sigma^2_{1}, \cdots, \sigma^2_{p})$ so that the elements of the response vector now become uncorrelated. Also, let us  group  $\bB$ as set of $P$ $M$-dimensional observations $\left(\bbeta_1, \bbeta_2, \hdots, \bbeta_P\right)$, and let us consider the predictor $\bx_n$ to be fixed, so that we just focus on the data distribution for now. These assumptions give us the following decomposition
\begin{align*}
 \prod_{n=1}^N  \sN_p( \by_{n} \mid \bB \bx_n, \ \bSigma) p\left(\bbeta_1, \bbeta_2, \hdots, \bbeta_P,\sigma^2_{1}, \cdots, \sigma^2_{P}\right)
\end{align*} 
 
By grouping the pairs $\left(\bbeta_p, \sigma^2_{p}\right)$  and then assuming conditional exchangeability of the parameter pairs, we can further decompose the model into
\begin{align*}
\prod_{n=1}^N  \sN_p(\by_{n} \mid  \bB \bx_n, \ \bSigma)=\prod_{n=1}^N \prod_{p=1}^P \sN(y_{np} \mid \bbeta^{\rm T}_p \bx_n, \ \sigma^2_p)p\left(\bbeta_p, \sigma^2_{p} \mid \xi\right) p(\xi)
\end{align*}

If we assume that the priors over the parameters  $\left(\bbeta_p, \sigma^2_{p}\right)$ to be uninformative, we further reduce the decomposition to 
\begin{align*}
\prod_{n=1}^N  \sN_p(\by_{n} \mid  \bB \bx_n, \ \bSigma)=\prod_{n=1}^N \prod_{p=1}^P \sN(y_{np} \mid \bbeta^{\rm T}_p \bx_n, \ \sigma^2_p)
\end{align*}
The above simplification implies that a multivariate regression model can be broken down into a set of $P$ multiple regression problems. For example, if we only consider one response variable $y \in \by$, we can then drop the subscript and obtain the following decomposition for a multiple linear regression model.
\begin{align*}
\prod_{n=1}^N\ p(y_{n}\mid \bx_n) \ &= \prod_{n=1}^N  \sN(y_{n} \mid\bbeta^{\rm T} \bx_n, \ \sigma^2)
\end{align*}

Note that the decomposition above is not exactly that of a hierarchical Bayesian regression model. In a hierarchical regression, the $N$ multivariate observation vector pairs $(\by_{n}, \bx_n)$ are partitioned into $J$ clusters and for each cluster $j$ we introduce a regression coefficient $\bbeta_{jp}$. Accordingly, the decomposition can be expressed as 
\begin{align*}
\prod_{n=1}^{N_j}  \prod_{j=1}^J \sN_p(\by_{n} \mid  \bB \bx_n, \ \bSigma)=\prod_{n=1}^{N_j}  \prod_{j=1}^J  \prod_{p=1}^P \sN(y_{njp} \mid \bbeta^{\rm T}_{jp} \bx_{nj}, \ \sigma^2_p) p\left(\bbeta_{jp}, \sigma^2_{p} \mid \xi \right) p(\xi)
\end{align*}

\subsection{Reduced Rank Regression (RRR)} \label{sec:RRR}

Our starting point is the decomposition given in Equation \ref{multivariate_normal} for the multivariate regression model - which is reproduced here.
\begin{align*}
 \prod_{n=1}^N  \sN_p( \by_{n} \mid \bB \bx_n, \ \bSigma) p(\bB,\bSigma)\prod_{n=1}^N\ p(\bx_n\mid \theta^{X})p(\theta^{X})
 \end{align*} 
 and the parameters that govern $\bx_n$  and  $\by_n$ to be independent, and 
\begin{align*}
\prod_{n=1}^N  \sN_p( \by_{n} \mid \bB \bx_n, \ \bSigma)
\end{align*} 

If we allow the rank of the regression coefficient matrix to be deficient such that $rank(\bB)=d \ < \ min(m, p)$, then the matrix  can be decomposed into two non-unique matrices as such $\overset{p\times m}{\mathbf{B}}=\overset{p\times d}{\mathbf{W}}\ \overset{d\times m}{\mathbf{D}}$. The decomposition results in what is known as Reduced Rank Regression (RRR) \citep{izenman1975} 
\begin{align*}
\prod_{n=1}^N  \sN_p( \by_{n} \mid \bW\bD\bx_n, \ \bSigma)
\end{align*}  

Now if set the predictor vector to be equal to the response vector then the resulting data distribution of the reduced rank regression can be expressed as 
\begin{align*}
\prod_{n=1}^N  \sN_p( \by_n \mid \bW\bD\by_n, \ \bSigma)
\end{align*}  

By letting ${\bz_n}=\overset{d\times m}{\mathbf{D}}\overset{m\times 1}{\by_n}$, we can now think of $\by_n$ as a message; ${\bz_n}$ as an encoded transformation of the message; and $\bW$ as the decoding transformation. The model can then be expressed as
\begin{align*}
\prod_{n=1}^N  \sN_p( \by_n \mid \bW\bz_n, \ \bSigma)
\end{align*}  

This representation allows us to see that several latent variable models are special cases of $RRR$ \citep{izenman2008}. 

\subsection{Principle Component Analysis (PCA)}

To obtain our first and simplest latent variable model, we specify  a distribution to the latent vector since the lower dimensional encoded message is unobservable. Assigning a standard Gaussian to the latent variables and restricting the covariance of the errors to be isotropic gives us probabilistic PCA \citep{tipping1999}. The model can be expressed as
\begin{align*}
\prod_{n=1}^N  \sN_p( \by_n \mid \bW\bz_n, \ \sigma^2\mathbf{I}) \mathcal{N}_d(\bz_n \mid \boldsymbol{0},\,\mathbf{I})
\end{align*}  
or as a generative model 
\begin{align*}
    \begin{split}
\by_n \mid  \bz_n &\sim\ \mathcal{N}_p(\bW\bz_n,\, \sigma^2\mathbf{I}) \\
\bz_n\ &\sim\ \mathcal{N}_d(\boldsymbol{0},\,\mathbf{I}) \ \quad \textit{for } n=1, \cdots,N
    \end{split}
\end{align*}

Standard PCA is obtained as the maximum likelihood solution if we let $\sigma^2 \rightarrow 0$ and constrain $\bW$ to be orthogonal.

\subsection{Factor Analysis (FA)}

If we relax the constraint of homogeneous variance (i.e. $\sigma^2$ is the same for all components of the vector $\by_n$), the covariance matrix can be specified as 
\begin{align*}
\bSigma=\Diag(\sigma^2)=\begin{bmatrix}
  \sigma^2_1 & 0  & \ldots & 0  \\
   0 & \sigma^2_2  & \ldots & 0  \\
 \vdots &\vdots &  \ddots  & \vdots \\
0 & \ldots  & 0 &\sigma^2_p
\end{bmatrix} 
\end{align*}
The result is a multidimensional factor analysis model with a diagonal covariance and $d$ latent factors.
\begin{align*}
\prod_{n=1}^N  \sN_p( \by_n \mid \bW\bz_n, \ \Diag(\sigma^2)) \mathcal{N}_d(\bz_n \mid \boldsymbol{0},\,\mathbf{I})
\end{align*}  
or as a generative model 
\begin{align}\label{eq:fa}
    \begin{split}
\by_n \mid  \bz_n &\sim\ \mathcal{N}_p(\bW\bz_n,\,\Diag(\sigma^2) ) \\
\bz_n\ &\sim\ \mathcal{N}_d(\boldsymbol{0},\,\mathbf{I}) \ \quad \textit{for } n=1, \cdots,N
    \end{split}
\end{align}
Note that the standard normal prior over the latent variables resolves  invariance in the scale and location of the latent space, but not in rotation.

\subsection{Independent Component Analysis (ICA)}

Independent Component Analysis \citep{Hyvarinen2015} is a widely used model for signal source separation. In order to model signals as latent variables, their distribution cannot be normal because the Gaussian is the most random of all distributions. In fact, according to \emph{Jayne’s principle of maximum entropy} \citep{jaynes2003}, a Gaussian random variable has maximum entropy (i.e.uncertainty) among all random variables with equal variance. In contrast, entropy tends to be small for spiked, or super-Gaussian distributions - distributions that characterize signals! 
So if we consider the \emph{generalized Gaussian distribution} as a prior over the latent variables, we obtain a highly peaked supergaussian distribution (if the shape parameter $\alpha$ set to be less than 2) that is appropriate for modeling signals that give rise to sparse data. Note that the version of the generalized Gaussian distribution we are considering is that of parametric symmetric distributions. The normal distribution is a special case obtained when $\alpha=2$ and the Laplace distribution is another special case when $\alpha=1$ \citep{gomez1998}. ICA can be thought of as a linear latent generative model, a type of a Factor Analysis model but with nongaussian priors over the latent variables (i.e. sources). The model has the following decomposition.
\begin{align*}
\prod_{n=1}^N  \sN_p( \by_n \mid \bW\bz_n, \ \Diag(\sigma^2)) \prod_{n=1}^N \prod_{d=1}^D \mathcal{GG}_d(z_{d,n} \mid\alpha_d) \quad \alpha_d < 2 
\end{align*}  
or as a generative model 
\begin{align*}
    \begin{split}
\by_n \mid  \bz_n&\sim\ \mathcal{N}_p(\bW\bz_n,\, \Diag(\sigma^2))\\
z_{d,n}\ &\sim\ \mathcal{GG}_d(\alpha_d) \quad \alpha_d < 2 \quad d=1:D; \ \quad n=1, \cdots,N
    \end{split}
\end{align*}
where $\mathcal{GG}$ is the generalized Gaussian distribution.

\subsection{Canonical Correlation Analysis (CCA)}

\citet{Hotelling1936} introduced Canonical Correlation Analysis (CCA) as a dimensionality reduction method that projects two sets of random variables (i.e. two data sets) into a shared subspace such that the projections are maximally correlated. Although almost always commonly thought as data reduction technique, CCA can indeed be given a latent variable interpretation as \citet{bach2005} have shown. More specifically, they showed that the equivalent probabilistic latent variable model can be formulated as   
\begin{align*}
    \begin{split}
\bz_n&\sim \mathcal{N}_d\left(\boldsymbol{0},\mathbf{I}\right)   \quad n=1, \cdots,N\\
\begin{bmatrix}
    {\by_n}^{(1)}\mid \bz_n\\
    {\by_n}^{(2)}\mid \bz_n
\end{bmatrix} &\sim \mathcal{N}_{(p1+p2)}\left(\begin{bmatrix}
    \mathbf{W}^{(1)}\\
    \mathbf{W}^{(2)} 
\end{bmatrix} \bz_n,\begin{bmatrix}
    \bSigma^{(1)} & \boldsymbol{0}\\
   \boldsymbol{0} & \bSigma^{(2)}
\end{bmatrix}\right) 
    \end{split}
\end{align*}
where ${\by_n}^{(1)}\in \mathbb{R}^{p_1}$ and ${\by_n}^{(2)}\in \mathbb{R}^{p_2}$, each denote a random vector of observations for Data set 1 and Data set 2, respectively. Note that by replacing the block diagonal covariance matrix of the data with a diagonal matrix, the probabilistic CCA model reduces to a standard factor analysis model!

They also show that since any covariance matrix of rank $D$ can be written as  the  following decomposition of the marginal covariance  in the marginal joint distribution of the data
\begin{align*}
\begin{bmatrix}
    {\by_n}^{(1)}\\
    {\by_n}^{(2)}
\end{bmatrix}&\sim \mathcal{N}\left(\boldsymbol{0},\begin{bmatrix}
    \bW^{(1)} {\bW^{(1)}}^\intercal +\bSigma^{(1)}   & \bW^{(1)} {\bW^{(2)}}^\intercal   \\
    \bW^{(2)} {\bW^{(1)}}^\intercal   & \bW^{(2)} {\bW^{(2)}}^\intercal  + \bSigma^{(2)}
\end{bmatrix}\right)
\end{align*}
it follows that the MLE solution for the joint covariance matrix is such that the cross-covariance matrix is of rank $D$ and consists of the canonical correlation directions that would be obtained in regular CCA.

\subsection{Inter-Battery Factor Analysis (IBFA)}

If we split the vector of latent variables $\bz_n$ into three subsets $\lbrace \bz_{n,0}, \bz_{n,1}, \bz_{n,1} \rbrace$  such that the latent vector $\bz_{n,0}$ captures the  common structure shared across both groups of variables, ${\by_n}^{(1)}$ and ${\by_n}^{(2)}$, while $\bz_{n,1}$ and  $\bz_{n,1}$  each captures the latent structure specific to each group, then CCA model can be modified to have an equivalent specification to the Inter-Battery Factor Analysis (IBFA) model \citep{tucker1958,browne1979}. 

\begin{align*}
    \begin{split}
\begin{bmatrix}
    \bz_{n,0}\\
    \bz_{n,1} \\
    \bz_{n,2}
\end{bmatrix}&\sim \mathcal{N}_d\left(\boldsymbol{0},\mathbf{I}\right)   \quad n=1, \cdots,N\\
\begin{bmatrix}
    \mathbf{x}^{(1)}\mid \bz_{n,0},\bz_{n,1}\\
    \mathbf{x}^{(2)}\mid \bz_{n,0},\bz_{n,2}
\end{bmatrix} &\sim \mathcal{N}\left(\begin{bmatrix}
    \bW^{(1,0)} & \bW^{(1,1)} &\boldsymbol{0}\\
    \bW^{(2,0)} & \boldsymbol{0} &\bW^{(2,2)}
\end{bmatrix} \begin{bmatrix}
    \bz_{n,0}\\
    \bz_{n,1} \\
    \bz_{n,2}
\end{bmatrix},\begin{bmatrix}
    \bSigma^{(1)} & \boldsymbol{0}\\
   \boldsymbol{0} & \bSigma^{(2)}
\end{bmatrix}\right)
    \end{split}
\end{align*}

Note that the above generative formulation makes it apparent that IBFA can be viewed as a special case of CCA, one obtained when we restrict the projection matrix $\mathbf{W}$ to have a the structure given above; namely
\begin{align*}
\mathbf{W}=
\begin{bmatrix}
    \mathbf{W}^{(1)}\\
    \mathbf{W}^{(2)} 
\end{bmatrix}=\begin{bmatrix}
    \bW^{(1,0)} & \bW^{(1,1)} &\boldsymbol{0}\\
    \bW^{(2,0)} & \boldsymbol{0} &\bW^{(2,2)}
\end{bmatrix}
\end{align*}

\subsection{Multi-Battery Factor Analysis (MBFA)}

The Inter-Battery Factor Analysis model can be readily extended to multiple groups of variables. So if we let ${\by_n}^{(g)} \in \mathbb{R}^{p_g}$ for $g=1, \cdots, G$ denote the groups of variables (i.e. data sets), the vector of observations can be stacked as $P= \sum_g^G p_g$ dimensional vector and the Multi-Battery Factor Analysis (MBFA) model \citep{browne1980} can be expressed as follows.
\begin{align*}
    \begin{split}
\bz_n&\sim \mathcal{N}_d\left(\boldsymbol{0},\mathbf{I}\right)   \quad n=1, \cdots,N\\
\begin{bmatrix}
    {\by_n}^{(1)}\mid \bz_n\\
    {\by_n}^{(2)}\mid \bz_n \\
    \vdots\\
    {\by_n}^{(G)}\mid \bz_n\\
\end{bmatrix} &\sim \mathcal{N}_{P}\left(\begin{bmatrix}
    \mathbf{W}^{(1)}\\
    \mathbf{W}^{(2)} \\
    \vdots\\
    \mathbf{W}^{(G)}
\end{bmatrix} \bz_n,\begin{bmatrix}
   \bSigma^{(1)} & 0  & \ldots & 0  \\
   0 & \bSigma^{(2)}  & \ldots & 0  \\
 \vdots &\vdots &  \ddots  & \vdots \\
0 & \ldots  & 0 &\bSigma^{(G)}
\end{bmatrix} \right)\\
    \end{split} 
\end{align*}

Again note that by replacing the block diagonal covariance matrix of the data with a diagonal matrix, MBFA reduces to a standard factor analysis model.

\subsection{Group Factor Analysis (GFA)}

\citet{klami2015} proposed Group Factor Analysis (GFA) as a generalization of multi-battery factor analysis (MBFA). The generative model does still have the same general structure of the preceding latent variable models

\begin{align*}
    \begin{split} 
\bz_n&\sim \mathcal{N}_d\left(\boldsymbol{0},\mathbf{I}\right)   \quad n=1, \cdots,N\\ 
\begin{bmatrix}
    {\by_n}^{(1)}\mid \bz_n\\
    {\by_n}^{(2)}\mid \bz_n \\
    \vdots\\
    {\by_n}^{(G)}\mid \bz_n\\
\end{bmatrix} &\sim \mathcal{N}_{P}\left(\begin{bmatrix}
    \mathbf{W}^{(1)}\\
    \mathbf{W}^{(2)} \\
    \vdots\\
    \mathbf{W}^{(G)}
\end{bmatrix} \bz_n,\begin{bmatrix}
   \sigma^{2}_1\mathbf{I} & 0  & \ldots & 0  \\
   0 & \sigma^{2}_2\mathbf{I}  & \ldots & 0  \\
 \vdots &\vdots &  \ddots  & \vdots \\
0 & \ldots  & 0 &\sigma^{2}_G\mathbf{I}
\end{bmatrix} \right)
    \end{split}  
\end{align*}
where  
\begin{align*}
\bW=
\begin{bmatrix}
    \mathbf{W}^{(1)}\\
    \mathbf{W}^{(2)} \\
    \vdots\\
    \mathbf{W}^{(G)}
\end{bmatrix} =\begin{bmatrix}
    \mathbf{w}^{(1)}_1 & \mathbf{w}^{(1)}_2 & \dots  & \mathbf{w}^{(1)}_D \\
    \mathbf{w}^{(2)}_1 & \mathbf{w}^{(2)}_2 & \dots  & \mathbf{w}^{(2)}_D \\
    \vdots & \vdots & \vdots & \vdots \\
    \mathbf{w}^{(G)}_1 &\mathbf{w}^{(G)}_2  & \dots  & \mathbf{w}^{(G)}_D 
\end{bmatrix}
\end{align*}

There are two differences however. First, GFA restricts the group specific error covariance to isotropic structure. That is, $\bSigma^{(g)}=\sigma^{2}_g\mathbf{I}$ for  $g=1, \cdots, G$. Second, in order to model the dependency structure between any subsets of groups and resolve indeterminacy issues, GFA imposes a structural sparsity prior over the columns of the projection matrix $\bW$. The Automatic Relevance Determination (ARD) prior \citep{mackay1995b} over the $G \times D$ group specific columns of the projection matrix $\bW$ can be specified as follows.
\begin{align*}
\begin{split}
\mathbf{w}^{(g)}_d&\sim \mathcal{N}\left(\boldsymbol{0},{(\alpha^{(g)}_{d})}^{-1}\mathbf{I}\right) \quad g=1, \cdots, G; \quad d=1,\cdots,D  \\
log(\boldsymbol{\alpha}) &= UV^\intercal + \mu_u \textbf{1}^\intercal + \textbf{1}\mu_v^\intercal   \\
p(U) &= \prod_g^G{\prod_r^R{\mathcal{N}(u_{g,r}|0, (\lambda=0.1)^{-1})}} \\
p(V) &= \prod_d^D{\prod_r^R{\mathcal{N}(v_{d,r}|0, (\lambda=0.1)^{-1})}}
\end{split}
\end{align*}
for $\boldsymbol{\alpha} \in \mathbb{R}^{G \times D}$, $U \in \mathbb{R}^{G \times R}$, and $V \in \mathbb{R}^{D \times R}$ and $R \ll min(G,D)$.

\subsection{Structural Equation Models (SEM)}

The preceding models can be extended by allowing the latent variables to regress on each other. The resulting general model is known as a Structural Equation Model (SEM) and was first introduced by \citet{Joreskog1970}. SEM models are constructed using algebraic equations, not by formulating a probabilistic statistical model as is done in Bayesian Graphical Models \citep{pearl2014} for example. 

A popular general formulation of SEMs is given by the LISREL model. The LISREL model consists of two parts. The first part is a structural component in which dependencies among the latent variables are created by regressing them on each other. The vector of latent variables we introduce into the model is divided into two groups of independent and dependent variable $\bz_n= (\bz_{n,1}^\intercal, \bz_{n,2}^\intercal)$ of dimensions $D_1$ and $D_2$, respectively, such that $D_1+D_2=D$. The second part consists of a measurement model that is similar in form to the factor analysis model (i.e. $\by_n =\bW\bz_n + \bepsilon_n$) but for which we also partition the $P$-dimensional observation vector (and correspondingly the error vector) into two groups $\by_n= (\by_{n,1}^\intercal, \by_{n,2}^\intercal)$ and $\bepsilon_n= ( \bepsilon_{n,1}^\intercal, \bepsilon_{n,2}^\intercal)$ such that $\by_{n,1}$ depends only on $\bz_{n,2}$ and $\by_{n,2}$ depends only on $\bz_{n,1}$. The dimension of two groups is $P_1$ and $P_2$, such that $P_1 + P_2=P$. The general algebraic form of the LISREL model is as follows.
\begin{align*}
    \begin{split} 
\begin{bmatrix}
    \by_{n,1} \\
    \by_{n,2}
\end{bmatrix}&=
\begin{bmatrix}
    \bW_1 & \boldsymbol{0}\\
    \boldsymbol{0} & \bW_2
\end{bmatrix}
\begin{bmatrix}
    \bz_{n,1} \\
    \bz_{n,2}
\end{bmatrix} +
\begin{bmatrix}
    \bepsilon_{n,1} \\
    \bepsilon_{n,2}
\end{bmatrix}\\
\bB\bz_{n,2}&= \bC\bz_{n,1}+ \bxi_n
    \end{split} 
\end{align*}
where it is assumed that B is non-singular and
\begin{align*}
    \begin{split} 
&\E(\by_n)= \bzero; \E(\bz_n)= \bzero\; \ \E(\bepsilon_n)= \bzero; \ \E(\bxi_n)= \bzero\\
&\Cov(\bepsilon_n)= \Psi; \ \Cov(\bz_{n,1})= \Phi_{z_1}; \ \Cov(\bxi_n)= \Phi_{\bxi}
   \end{split} 
\end{align*}
and the covariance matrices are diagonal. The regressions imply the following covariance matrix $\bSigma_{\by\by}$ for the observations.

\begin{align*}
\begin{bmatrix}
\bW_1 \Phi_{z_1} \bW_1^\intercal + \Psi_1 & 
\bW_1 \Phi_{z_1} \bC^\intercal{(\bB^\intercal)}^{-1}  \bW_2^\intercal\\
\bW_2\Phi_{z_1} \bC^\intercal{(\bB^\intercal)}^{-1} \bW_1^\intercal & 
\bW_2 (\bB^{-1}\bC^\intercal\Phi_{z_1}\bC{(\bB^\intercal)}^{-1} + \bB^{-1}\Phi_{\bxi}{(\bB^\intercal)}^{-1}) \bW_2^\intercal  +\Psi_2
\end{bmatrix}
\end{align*}

Notice that how the implied covariance structure of the data is determined by the coefficient matrices ( i.e. $\bB$, $\bC$, and $\bW$) and the covariance matrices (i.e. $\Psi$,  $\Phi_{z_1}$ and $\Phi_{\bxi}$). More importantly, notice also that an SEM model implicitly assumes that the relationships between the observations are linear since covariances are measures of linear association. Another aspect of SEM models that is worth pointing out is that although we can assess whether a particular implied decomposition is consistent with the data, we cannot determine whether it is the only decomposition that is equally consistent with the data. That is, the same data can be generated given different latent structural relations \citep{bartholomew2011}. 

\paragraph{Generalized Structured Component Analysis (GSCA)} A generalization of the SEM framework has recently been proposed by \citet{hwang2004}. The goal of the generalization is to unify two distinct approaches to SEM; namely the factor-based and component-based (i.e. partial least squares). The  Generalized Structured Component Analysis (GSCA) model they propose has the following structural formulation. 
\begin{align}\label{GSCA}
    \begin{split} 
    \begin{bmatrix}
    \mathbf{I}\\
    \mathbf{W} 
\end{bmatrix}\by_n=
\begin{bmatrix}
    \mathbf{C}\\
    \mathbf{B}
\end{bmatrix}\bW\by_n +
\begin{bmatrix}
    \bepsilon_n \\
    \bxi_n
\end{bmatrix}    
    \end{split} 
\end{align}
where the latent variables are obtained as component scores using what the authors refer to as weighted relation model $\bz_n=\bW\by_n$. The assumption regarding the means and covariance of the variables are identical to the LISERL model. It is interesting to note that the weighted relation model equation is no different conceptually than the encoding transformation that \citet{izenman2008} used to construct latent variable models from the reduced rank regression framework ( Section \ref{sec:RRR}). 

We argue that by relying on complicated algebraic equations to induce linear dependencies in the latent structure, the SEM framework risks losing statistical rigor especially when modeling complex phenomena. Real data, most often than not, is noisy and characterized by complex nonlinear dependencies. We think that we have a better and more statistically rigorous alternative in the probabilistic generative perspective that approaches the modeling endeavor by first proposing a probabilistic model for all the observed variables and  any latent variables that we wish to incorporate in the model. As part of the probabilistic model, we can also specify functional relations between the variables in order to capture any nonlinear dependencies in the data or to construct a rich latent space with complex structure. For example, a recent method, the Inverse Autoregressive Flow (IAF)\citep{kingma2016}, has been proposed in the generative modeling literature that enables us to create models with rich latent structure. The key idea behind the IAF is that any multivariate normal variable with diagonal covariance can be transformed into a multivariate normal with linear dependencies (i.e. full covariance matrix) by expressing it as an autoregressive model
\begin{align*}
z_d = \mu_d(\bz_{< d})+ \sigma_d(\bz_{ < d})\ast\epsilon
\end{align*}
where $\mu_d$ and $\sigma_d$ are the conditional mean vector and covariance matrix of $z_d$ on the other elements of the latent vector $\bz_n=(z_1, z_2, \cdots, z_d, \cdots, z_D)$. Accordingly, a general SEM model can be conceptually viewed as the following generative model.
\begin{align*}
\begin{split} 
\by_n \mid \bz_n &\sim \mathcal{N}_p\Big(\bW\bz_n, \Diag(\sigma^2)\Big)\\
\bz_n &\sim \mathcal{N}_r(\boldsymbol{0},\bSigma_z)
\end{split} 
\end{align*}

By including structural latent regressions, a specific SEM model would then correspond to a particular structure of the covariance matrix $\bSigma_z$. In the next section, we briefly describe Deep Latent Gaussian Models (DLGMs) as an alternative general framework to SEM, one that is more statistically disciplined and potentially much more powerful, especially when large amounts of data is available.  

\subsection{Deep Latent Gaussian Models (DLGM)}\label{model:DLGM}

A Deep Latent Gaussian Model (DLGM) \citep{rezende2014} is a directed graphical model with Gaussian latent variables that are stacked in $L$ stochastic layers and that are transformed at each layer by a differentiable nonlinear function parameterized by a neural network $\mathbf{NN}(\bz;\btheta)$ consisting of $K^{(l)}$ deterministic hidden layers. 
Whereas the latent variables in the first DLGMs such as the Variational Autoencoder (VAE) \citep{Kingma2013} are assumed to be independent, recent extensions \citep{kingma2016, maaloe2016} have introduced a deep latent hierarchy that endows the latent space with complex rich structure that can capture complex patterns of dependencies between the latent variables. A general DLGM model has the following generative formulation.

\begin{align*}
    \begin{split} 
     \mathbf{z}_n^{(L)}\ &\sim\ \mathcal{N}_{d_{(L)}}\left(\boldsymbol{0},\, \mathbf{I}\right)  \quad n=1, \cdots,N \\
    \bz_n^{(l)}\ &\sim\ \mathcal{N}_{d_{(l)}}\Big(\bz_n^{(l)} \mid \mathbf{NN}^{(l)}(\bz_n^{(l+1)};\btheta^{(l)}),\, \Sigma^{(l)} \Big) \quad l=1, \cdots,L-1\\
\by_n \mid  \bz_n &\sim\ \mathbf{Expon}\Big(\by_n \mid \mathbf{NN}^{(0)}(\bz_n^{(1)};\btheta^{(0)})\Big)
    \end{split} 
\end{align*}
where 
\begin{align*}
    \begin{split} 
\mathbf{NN}(\bz;\theta) &= h_K \circ h_{K-1} \circ \ldots \circ h_0(\mathbf{z}) \\
 \quad h_k(\by) &= \sigma_k(\mathbf{W}^{(k)} \by+\mathbf{b}^{(k)})\\  \btheta &= \{ (\mathbf{W}^{(k)}, \mathbf{b}^{(k)}) \}_{k=0}^K
    \end{split} 
\end{align*}
\\
Such that the observation model $\mathbf{Expon}$ refers to the  exponential family of models and the neural network $\mathbf{NN}(\bz;\btheta)$ consists of the composition of $K$ non-linear transformation functions, each of which is represented by a hidden layer $h_k(\by)$. The function $\sigma_k$ is an element-wise \textit{activation function} such as the sigmoid function or the rectified linear unit (ReLU). The constant $d_{(l)}$ is the number of latent variables at the $l$ layer so that the total number of latent variables equals $d=\sum_{l=1}^{L}d_{(l)}$

A Deep Latent Gaussian Model is considered a deep model in two distinct senses. First, the latent variables that make up the stochastic layers can be stacked in a deep hierarchy. Second, the neural networks that make up the deterministic nonlinear transformations can  be made up of several hidden layers. In contrast,  a nonlinear regression model is a \textit{shallow nonlinear} model in the sense that it uses only one  non-linear transformation while a factor analysis model is a \textit{shallow linear} model in the sense that the latent variables $\bz_n$ are linearly projected by a matrix $\bW$ to a higher dimensional vector.

\vskip 0.2in
\bibliography{farouni17}

\begin{thebibliography}{69}
\providecommand{\natexlab}[1]{#1}
\providecommand{\url}[1]{\texttt{#1}}
\expandafter\ifx\csname urlstyle\endcsname\relax
  \providecommand{\doi}[1]{doi: #1}\else
  \providecommand{\doi}{doi: \begingroup \urlstyle{rm}\Url}\fi

\bibitem[Advani and Ganguli(2016)]{Advani2016}
Madhu Advani and Surya Ganguli.
\newblock {Statistical Mechanics of Optimal Convex Inference in High
  Dimensions}.
\newblock \emph{Physical Review X}, 6\penalty0 (3):\penalty0 031034, 2016.

\bibitem[Airy(1861)]{airy1861}
George~Biddell Airy.
\newblock \emph{On the algebraical and numerical theory of errors of
  observations and the combination of observations}.
\newblock Macmillan\&Company, 1861.

\bibitem[Athreya and Lahiri(2006)]{athreya2006}
Krishna~B Athreya and Soumendra~N Lahiri.
\newblock \emph{Measure theory and probability theory}.
\newblock Springer Science \& Business Media, 2006.

\bibitem[Bach and Jordan(2005)]{bach2005}
Francis~R Bach and Michael~I Jordan.
\newblock {A probabilistic interpretation of canonical correlation analysis}.
\newblock techreport 688, Department of Statistics, University of California,
  Berkeley, Berkeley, 2005.

\bibitem[Banerjee et~al.(2014)Banerjee, Carlin, and Gelfand]{banerjee2014}
Sudipto Banerjee, Bradley~P Carlin, and Alan~E Gelfand.
\newblock \emph{Hierarchical modeling and analysis for spatial data}.
\newblock Crc Press, 2014.

\bibitem[Bartholomew et~al.(2011)Bartholomew, Knott, and
  Moustaki]{bartholomew2011}
David~J Bartholomew, Martin Knott, and Irini Moustaki.
\newblock \emph{Latent variable models and factor analysis: A unified
  approach}, volume 904.
\newblock John Wiley \& Sons, 2011.

\bibitem[Birnbaum(1968)]{Birnbaum1968}
A.~Birnbaum.
\newblock {Some latent trait models and their use in inferring an examinee's
  ability.}, 1968.

\bibitem[Bishop et~al.(1998)Bishop, Svens{\'e}n, and Williams]{bishop1998}
Christopher~M Bishop, Markus Svens{\'e}n, and Christopher~KI Williams.
\newblock Gtm{:} the generative topographic mapping.
\newblock \emph{Neural computation}, 10\penalty0 (1):\penalty0 215--234, 1998.

\bibitem[Blei et~al.(2016)Blei, Kucukelbir, and McAuliffe]{blei2016}
David~M Blei, Alp Kucukelbir, and Jon~D McAuliffe.
\newblock Variational inference: A review for statisticians.
\newblock \emph{Journal of the American Statistical Association (to appear).
  arXiv preprint arXiv:1601.00670}, 2016.

\bibitem[Bollen(2002)]{Bollen2002}
Kenneth~A Bollen.
\newblock Latent variables in psychology and the social sciences.
\newblock \emph{Annual review of psychology}, 53\penalty0 (1):\penalty0
  605--634, 2002.

\bibitem[Browne(1979)]{browne1979}
Michael~W Browne.
\newblock The maximum-likelihood solution in inter-battery factor analysis.
\newblock \emph{British Journal of Mathematical and Statistical Psychology},
  32\penalty0 (1):\penalty0 75--86, 1979.

\bibitem[Browne(1980)]{browne1980}
Michael~W Browne.
\newblock Factor analysis of multiple batteries by maximum likelihood.
\newblock \emph{British Journal of Mathematical and Statistical Psychology},
  33\penalty0 (2):\penalty0 184--199, 1980.

\bibitem[Chung et~al.(2015)Chung, Kastner, Dinh, Goel, Courville, and
  Bengio]{chung2015}
Junyoung Chung, Kyle Kastner, Laurent Dinh, Kratarth Goel, Aaron~C Courville,
  and Yoshua Bengio.
\newblock A recurrent latent variable model for sequential data.
\newblock In \emph{Advances in neural information processing systems}, pages
  2980--2988, 2015.

\bibitem[Diggle and Gratton(1984)]{diggle1984}
Peter~J Diggle and Richard~J Gratton.
\newblock Monte carlo methods of inference for implicit statistical models.
\newblock \emph{Journal of the Royal Statistical Society. Series B
  (Methodological)}, pages 193--227, 1984.

\bibitem[Durrett(2010)]{durrett2010}
Rick Durrett.
\newblock \emph{Probability: theory and examples}.
\newblock Cambridge university press, 2010.

\bibitem[Gelman et~al.(2014)Gelman, Carlin, Stern, and Rubin]{gelman2014}
Andrew Gelman, John~B Carlin, Hal~S Stern, and Donald~B Rubin.
\newblock \emph{Bayesian data analysis}, volume~2.
\newblock Chapman \& Hall/CRC Boca Raton, FL, USA, 2014.

\bibitem[Gemici et~al.(2017)Gemici, Hung, Santoro, Wayne, Mohamed, Rezende,
  Amos, and Lillicrap]{gemici2017}
Mevlana Gemici, Chia-Chun Hung, Adam Santoro, Greg Wayne, Shakir Mohamed,
  Danilo~J Rezende, David Amos, and Timothy Lillicrap.
\newblock Generative temporal models with memory.
\newblock \emph{arXiv preprint arXiv:1702.04649}, 2017.

\bibitem[Ghosh and Ramamoorthi(2006)]{Ghosh2006}
J.K. Ghosh and R.V. Ramamoorthi.
\newblock \emph{{Bayesian Nonparametrics}}.
\newblock Springer Science {\&} Business Media, 2006.

\bibitem[G{\'o}mez et~al.(1998)G{\'o}mez, Gomez-Viilegas, and Marin]{gomez1998}
E~G{\'o}mez, MA~Gomez-Viilegas, and JM~Marin.
\newblock A multivariate generalization of the power exponential family of
  distributions.
\newblock \emph{Communications in Statistics-Theory and Methods}, 27\penalty0
  (3):\penalty0 589--600, 1998.

\bibitem[Goodfellow et~al.(2014)Goodfellow, Pouget-Abadie, Mirza, Xu,
  Warde-Farley, Ozair, Courville, and Bengio]{goodfellow2014}
Ian Goodfellow, Jean Pouget-Abadie, Mehdi Mirza, Bing Xu, David Warde-Farley,
  Sherjil Ozair, Aaron Courville, and Yoshua Bengio.
\newblock Generative adversarial nets.
\newblock In \emph{Advances in neural information processing systems}, pages
  2672--2680, 2014.

\bibitem[Gould(1996)]{gould1996}
Stephen~Jay Gould.
\newblock \emph{The mismeasure of man}.
\newblock WW Norton \& Company, 1996.

\bibitem[H{\"a}gglund(2001)]{Hagglund2001}
G~H{\"a}gglund.
\newblock Milestones in the history of factor analysis.
\newblock In Robert Cudeck, K.~G. J{\"o}reskog, and Dag S{\"o}rbom, editors,
  \emph{Structural equation modeling, present and future: A festschrift in
  honor of Karl J{\"o}reskog}, pages 11--38. Scientific Software International,
  Lincolnwood, IL, 2001.

\bibitem[Hilborn and Mangel(1997)]{hilborn1997}
Ray Hilborn and Marc Mangel.
\newblock \emph{The ecological detective: confronting models with data},
  volume~28.
\newblock Princeton University Press, 1997.

\bibitem[Hotelling(1933)]{Hotelling1933}
Harold Hotelling.
\newblock {Analysis of a complex of statistical variables into principal
  components.}
\newblock \emph{Journal of Educational Psychology}, 24\penalty0 (6):\penalty0
  417--441, 1933.

\bibitem[Hotelling(1936)]{Hotelling1936}
Harold Hotelling.
\newblock {Relations Between Two Sets of Variates}.
\newblock \emph{Biometrika}, 28\penalty0 (3/4):\penalty0 321, dec 1936.

\bibitem[Hwang and Takane(2004)]{hwang2004}
Heungsun Hwang and Yoshio Takane.
\newblock Generalized structured component analysis.
\newblock \emph{Psychometrika}, 69\penalty0 (1):\penalty0 81--99, 2004.

\bibitem[Hwang and Takane(2014)]{hwang2014}
Heungsun Hwang and Yoshio Takane.
\newblock Generalized structured component analysis: A component-based approach
  to structural equation modeling, 2014.

\bibitem[Hyv{\"{a}}rinen(2015)]{Hyvarinen2015}
Aapo Hyv{\"{a}}rinen.
\newblock {A unified probabilistic model for independent and principal
  component analysis}.
\newblock \emph{Advances in Independent Component Analysis and Learning
  Machines}, pages 1--9, 2015.

\bibitem[Izenman(1975)]{izenman1975}
Alan~Julian Izenman.
\newblock Reduced-rank regression for the multivariate linear model.
\newblock \emph{Journal of multivariate analysis}, 5\penalty0 (2):\penalty0
  248--264, 1975.

\bibitem[Izenman(2008)]{izenman2008}
Alan~Julian Izenman.
\newblock \emph{Modern multivariate statistical techniques}.
\newblock Springer, 2008.

\bibitem[Jaynes(2003)]{jaynes2003}
Edwin~T Jaynes.
\newblock \emph{Probability theory: The logic of science}.
\newblock Cambridge university press, 2003.

\bibitem[Jebara(2012)]{jebara2012}
Tony Jebara.
\newblock \emph{Machine learning: discriminative and generative}, volume 755.
\newblock Springer Science \& Business Media, 2012.

\bibitem[J{\"o}reskog(1970)]{Joreskog1970}
K.~G. J{\"o}reskog.
\newblock {A general method for analysis of covariance structures}.
\newblock \emph{Biometrika}, 57\penalty0 (2):\penalty0 239--251, 1970.

\bibitem[Kallenberg(2006)]{kallenberg2006}
Olav Kallenberg.
\newblock \emph{Probabilistic symmetries and invariance principles}.
\newblock Springer Science \& Business Media, 2006.

\bibitem[Kant and Guyer(1998)]{kant1998}
Immanuel Kant and Paul Guyer.
\newblock \emph{Critique of pure reason}.
\newblock Cambridge University Press, 1998.

\bibitem[Kennedy and O'Hagan(2001)]{Kennedy2001}
Marc~C. Kennedy and Anthony O'Hagan.
\newblock Bayesian calibration of computer models.
\newblock \emph{Journal of the Royal Statistical Society: Series B (Statistical
  Methodology)}, 63, 2001.

\bibitem[Kingma and Welling(2013)]{Kingma2013}
Diederik~P Kingma and Max Welling.
\newblock {Auto-Encoding Variational Bayes}.
\newblock \emph{ICLR}, \penalty0 (Ml):\penalty0 1--14, dec 2013.

\bibitem[Kingma et~al.(2016)Kingma, Salimans, and Welling]{kingma2016}
Diederik~P Kingma, Tim Salimans, and Max Welling.
\newblock Improving variational inference with inverse autoregressive flow.
\newblock \emph{arXiv preprint arXiv:1606.04934v2}, 2016.

\bibitem[Klami et~al.(2015)Klami, Virtanen, Lepp{\"a}aho, and Kaski]{klami2015}
Arto Klami, Seppo Virtanen, Eemeli Lepp{\"a}aho, and Samuel Kaski.
\newblock Group factor analysis.
\newblock \emph{IEEE transactions on neural networks and learning systems},
  26\penalty0 (9):\penalty0 2136--2147, 2015.

\bibitem[Lake et~al.(2015)Lake, Salakhutdinov, and Tenenbaum]{lake2015}
Brenden~M Lake, Ruslan Salakhutdinov, and Joshua~B Tenenbaum.
\newblock Human-level concept learning through probabilistic program induction.
\newblock \emph{Science}, 350\penalty0 (6266):\penalty0 1332--1338, 2015.

\bibitem[Lauritzen(1996)]{lauritzen1996}
Steffen~L Lauritzen.
\newblock \emph{Graphical models}, volume~17.
\newblock Clarendon Press, 1996.

\bibitem[Lazarsfeld(1950)]{lazarsfeld1950}
Paul~F Lazarsfeld.
\newblock The logical and mathematical foundation of latent structure analysis.
\newblock \emph{Measurement and prediction}, 4:\penalty0 362--412, 1950.

\bibitem[Lord(1952)]{lord1952}
Frederic~M Lord.
\newblock The relation of test score to the trait underlying the test.
\newblock \emph{ETS Research Report Series}, 1952\penalty0 (2):\penalty0
  517--549, 1952.

\bibitem[Maal{\o}e et~al.(2016)Maal{\o}e, S{\o}nderby, S{\o}nderby, and
  Winther]{maaloe2016}
Lars Maal{\o}e, Casper~Kaae S{\o}nderby, S{\o}ren~Kaae S{\o}nderby, and Ole
  Winther.
\newblock Auxiliary deep generative models.
\newblock \emph{arXiv preprint arXiv:1602.05473}, 2016.

\bibitem[MacKay(1995{\natexlab{a}})]{mackay1995a}
David~JC MacKay.
\newblock Bayesian neural networks and density networks.
\newblock \emph{Nuclear Instruments and Methods in Physics Research Section A:
  Accelerators, Spectrometers, Detectors and Associated Equipment},
  354\penalty0 (1):\penalty0 73--80, 1995{\natexlab{a}}.

\bibitem[MacKay(1995{\natexlab{b}})]{mackay1995b}
David~JC MacKay.
\newblock Probable networks and plausible predictions - a review of practical
  bayesian methods for supervised neural networks.
\newblock \emph{Network: Computation in Neural Systems}, 6\penalty0
  (3):\penalty0 469--505, 1995{\natexlab{b}}.

\bibitem[Mansinghka et~al.(2016)Mansinghka, Shafto, Jonas, Petschulat, Gasner,
  and Tenenbaum]{mansinghka2016}
Vikash Mansinghka, Patrick Shafto, Eric Jonas, Cap Petschulat, Max Gasner, and
  Joshua~B Tenenbaum.
\newblock Crosscat: A fully bayesian nonparametric method for analyzing
  heterogeneous, high dimensional data.
\newblock \emph{Journal of Machine Learning Research}, 17:\penalty0 1--49,
  2016.

\bibitem[Muth{\'e}n(2002)]{muthen2002}
Bengt~O Muth{\'e}n.
\newblock Beyond sem: General latent variable modeling.
\newblock \emph{Behaviormetrika}, 29\penalty0 (1):\penalty0 81--117, 2002.

\bibitem[Pearl(2014)]{pearl2014}
Judea Pearl.
\newblock \emph{Probabilistic reasoning in intelligent systems: networks of
  plausible inference}.
\newblock Morgan Kaufmann, 2014.

\bibitem[Pearson(1901)]{Pearson1901}
Karl Pearson.
\newblock {LIII. On lines and planes of closest fit to systems of points in
  space}.
\newblock \emph{Philosophical Magazine Series 6}, 2\penalty0 (11):\penalty0
  559--572, nov 1901.

\bibitem[Pritchard et~al.(1999)Pritchard, Seielstad, Perez-Lezaun, and
  Feldman]{pritchard1999}
Jonathan~K Pritchard, Mark~T Seielstad, Anna Perez-Lezaun, and Marcus~W
  Feldman.
\newblock Population growth of human y chromosomes: a study of y chromosome
  microsatellites.
\newblock \emph{Molecular biology and evolution}, 16\penalty0 (12):\penalty0
  1791--1798, 1999.

\bibitem[Rabe-Hesketh et~al.(2004)Rabe-Hesketh, Skrondal, and
  Pickles]{rabe2004}
Sophia Rabe-Hesketh, Anders Skrondal, and Andrew Pickles.
\newblock Generalized multilevel structural equation modeling.
\newblock \emph{Psychometrika}, 69\penalty0 (2):\penalty0 167--190, 2004.

\bibitem[Ranganath et~al.(2015)Ranganath, Tang, Charlin, and
  Blei]{ranganath2015}
Rajesh Ranganath, Linpeng Tang, Laurent Charlin, and David~M Blei.
\newblock Deep exponential families.
\newblock In \emph{AISTATS}, 2015.

\bibitem[Rezende et~al.(2014)Rezende, Mohamed, and Wierstra]{rezende2014}
Danilo~Jimenez Rezende, Shakir Mohamed, and Daan Wierstra.
\newblock Stochastic backpropagation and approximate inference in deep
  generative models.
\newblock \emph{arXiv preprint arXiv:1401.4082}, 2014.

\bibitem[Rezende et~al.(2016)Rezende, Mohamed, Danihelka, Gregor, and
  Wierstra]{rezende2016}
Danilo~Jimenez Rezende, Shakir Mohamed, Ivo Danihelka, Karol Gregor, and Daan
  Wierstra.
\newblock One-shot generalization in deep generative models.
\newblock \emph{arXiv preprint arXiv:1603.05106}, 2016.

\bibitem[Roweis and Ghahramani(1999)]{Roweis1999}
Sam Roweis and Zoubin Ghahramani.
\newblock {A Unifying Review of Linear Gaussian Models}.
\newblock \emph{Neural Computation}, 1999.

\bibitem[Searle et~al.(2009)Searle, Casella, and McCulloch]{searle2009}
Shayle~R Searle, George Casella, and Charles~E McCulloch.
\newblock \emph{Variance components}, volume 391.
\newblock John Wiley \& Sons, 2009.

\bibitem[Skrondal and Rabe-Hesketh(2004)]{Skrondal2004}
Anders Skrondal and Sophia Rabe-Hesketh.
\newblock \emph{{Generalized Latent Variable Modeling}}, volume 20041561 of
  \emph{C{\&}H/CRC Monographs on Statistics {\&} Applied Probability}.
\newblock Chapman and Hall/CRC, may 2004.

\bibitem[Sobel(1994)]{sobel1994}
Michael~E Sobel.
\newblock Causal inference in latent variable models.
\newblock 1994.

\bibitem[Spearman(1904)]{Spearman1904}
C~Spearman.
\newblock {"General Intelligence," Objectively Determined and Measured}.
\newblock \emph{The American Journal of Psychology}, 15\penalty0 (2):\penalty0
  201, apr 1904.

\bibitem[Tavar{\'e} et~al.(1997)Tavar{\'e}, Balding, Griffiths, and
  Donnelly]{tavare1997}
Simon Tavar{\'e}, David~J Balding, Robert~C Griffiths, and Peter Donnelly.
\newblock Inferring coalescence times from dna sequence data.
\newblock \emph{Genetics}, 145\penalty0 (2):\penalty0 505--518, 1997.

\bibitem[Thomson(1916)]{Thomson1916}
Godfrey~H. Thomson.
\newblock {A HIERARCHY WITHOUT A GENERAL FACTOR}.
\newblock \emph{British Journal of Psychology, 1904-1920}, 8\penalty0
  (3):\penalty0 271--281, sep 1916.

\bibitem[Thurstone and L.L.(1947)]{Thurstone1947}
Thurstone and L.L.
\newblock {Multiple Factor Analysis.}, 1947.

\bibitem[Tipping and Bishop(1999)]{tipping1999}
Michael~E. Tipping and Christopher~M. Bishop.
\newblock {Probabilistic Principal Component Analysis}.
\newblock \emph{Journal of the Royal Statistical Society: Series B (Statistical
  Methodology)}, 61\penalty0 (3):\penalty0 611--622, aug 1999.

\bibitem[Tobin(1958)]{tobin1958}
James Tobin.
\newblock Estimation of relationships for limited dependent variables.
\newblock \emph{Econometrica: journal of the Econometric Society}, pages
  24--36, 1958.

\bibitem[Tornio et~al.()Tornio, Honkela, and Karhunen]{tornio2007}
Matti Tornio, Antti Honkela, and Juha Karhunen.
\newblock Time series prediction with variational bayesian nonlinear
  state-space models.

\bibitem[Tran et~al.(2017)Tran, Ranganath, and Blei]{tran2017}
Dustin Tran, Rajesh Ranganath, and David~M Blei.
\newblock Deep and hierarchical implicit models.
\newblock \emph{arXiv preprint arXiv:1702.08896}, 2017.

\bibitem[Tucker(1958)]{tucker1958}
Ledyard~R Tucker.
\newblock An inter-battery method of factor analysis.
\newblock \emph{Psychometrika}, 23\penalty0 (2):\penalty0 111--136, 1958.

\bibitem[Wright(1921)]{wright1921}
Sewall Wright.
\newblock Correlation and causation.
\newblock \emph{Journal of agricultural research}, 20\penalty0 (7):\penalty0
  557--585, 1921.

\end{thebibliography}

\end{document}